\documentclass[10pt,twocolumn,letterpaper]{article}

\usepackage[pagenumbers]{cvpr} 

\usepackage{url}
\usepackage{multirow}
\usepackage{graphicx}
\usepackage{float}
\usepackage{arydshln}
\usepackage{amsmath}
\usepackage{booktabs}
\usepackage{wrapfig}
\usepackage[ruled,vlined]{algorithm2e}
\usepackage[dvipsnames]{xcolor}

\definecolor{cvprblue}{rgb}{0.21,0.49,0.74}
\usepackage[pagebackref,breaklinks,colorlinks,allcolors=cvprblue]{hyperref}

\def\paperID{} 
\def\confName{CVPR}
\def\confYear{2026}

\title{Diffusion-SDPO: Safeguarded Direct Preference Optimization for Diffusion Models}

\author{
  Minghao Fu$^{1,2,3,}$\thanks{Work done during the internship at Alibaba International Digital Commerce Group.} \quad
  Guo-Hua Wang$^{3,}$\thanks{G. Wang is the corresponding author.} \quad
  Tianyu Cui$^{3}$ \quad
  Qing-Guo Chen$^{3}$ \\
  {Zhao Xu$^{3}$} \quad
  Weihua Luo$^{3}$ \quad
  Kaifu Zhang$^{3}$ \\[0.5em]
  $^1$School of Artificial Intelligence, Nanjing University \\
  $^2$National Key Laboratory for Novel Software Technology, Nanjing University \\
  $^3$Alibaba International Digital Commerce Group \\[0.5em]
  \texttt{fumh@lamda.nju.edu.cn\quad wangguohua@alibaba-inc.com}
}

\begin{document}

\maketitle

\begin{abstract}
Text-to-image diffusion models deliver high-quality images, yet aligning them with human preferences remains challenging. We revisit diffusion-based Direct Preference Optimization (DPO) for these models and identify a critical pathology: enlarging the preference margin does not necessarily improve generation quality. In particular, the standard Diffusion-DPO objective can increase the reconstruction error of both winner and loser branches. Consequently, degradation of the less-preferred outputs can become sufficiently severe that the preferred branch is also adversely affected even as the margin grows. To address this, we introduce Diffusion-SDPO, a safeguarded update rule that preserves the winner by adaptively scaling the loser gradient according to its alignment with the winner gradient. A first-order analysis yields a closed-form scaling coefficient that guarantees the error of the preferred output is non-increasing at each optimization step. Our method is simple, model-agnostic, broadly compatible with existing DPO-style alignment frameworks and adds only marginal computational overhead. Across standard text-to-image benchmarks, Diffusion-SDPO delivers consistent gains over preference-learning baselines on automated preference, aesthetic, and prompt alignment metrics. Code is publicly available at \url{https://github.com/AIDC-AI/Diffusion-SDPO}.
\end{abstract}

\section{Introduction}

\begin{figure*}
  \centering
  \includegraphics[width=.8\linewidth]{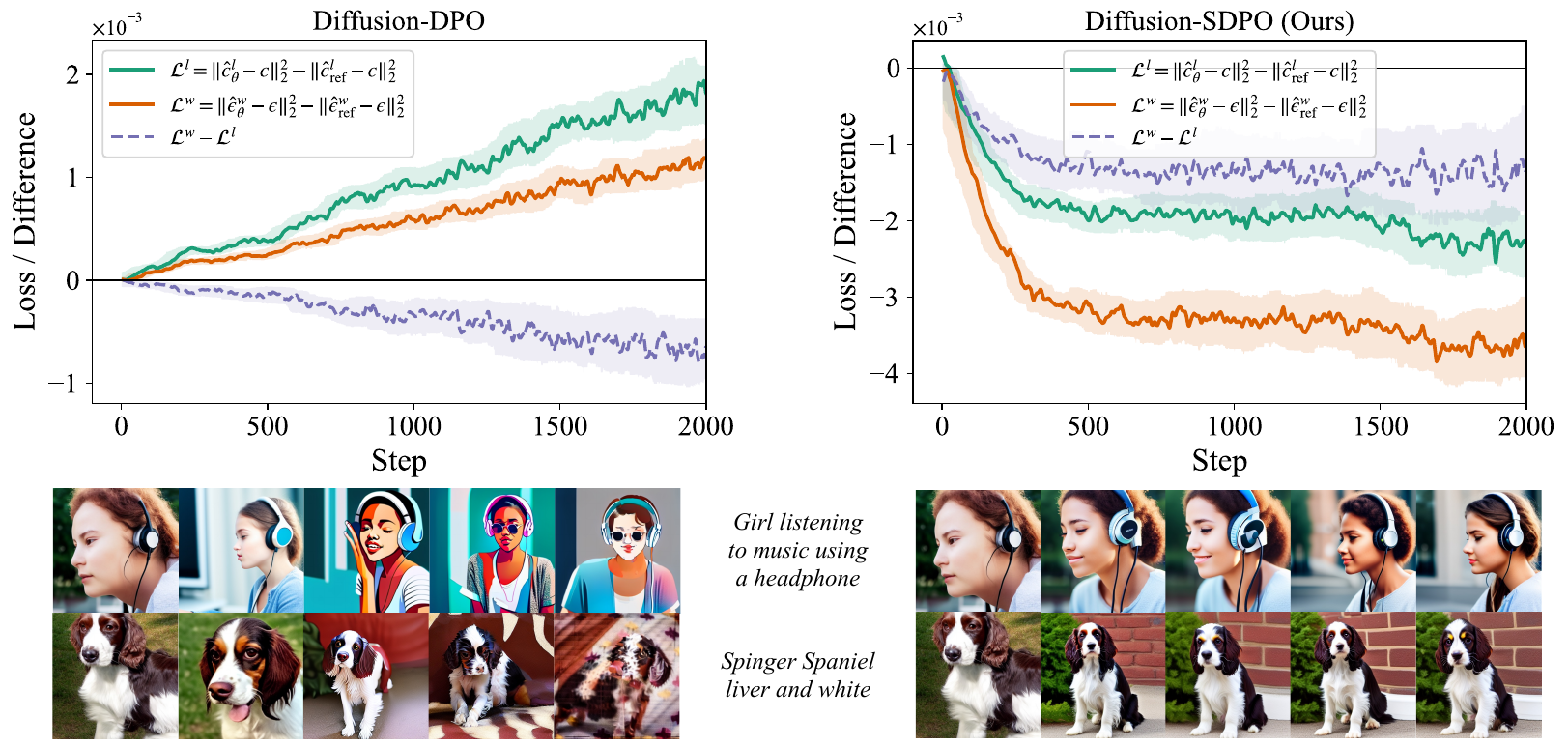}
  \caption{Training dynamics of preference losses during DPO finetuning \textit{without (left)} and \textit{with (right)} our safe-$\lambda$ mechanism on SD 1.5~\cite{sd15}. Images beneath the plots illustrate samples generated at training steps $\{0,500,1000,1500,2000\}$.}
  \label{fig:insight_combined}
\end{figure*}

Text-to-image diffusion models~\cite{diffusion_survey} have achieved remarkable success in generating diverse and high-quality images~\cite{flux,google2025_nano_banana}. However, aligning these powerful generative models with nuanced human preferences remains a critical challenge. Recent approaches have begun to incorporate human feedback~\cite{rlhf} into diffusion model training, drawing inspiration from alignment techniques used in large language models. In particular, Direct Preference Optimization (DPO)~\cite{llm_dpo} has emerged as a promising alternative to reinforcement learning for finetuning on human preferences. DPO directly optimizes the model on pairwise human comparisons (winner vs. loser outputs), and has been successfully adapted to text-to-image diffusion models in methods~\cite{diffusion-dpo,mapo,dspo,dmpo,diffusion-kto} to improve visual appeal and prompt alignment. Despite these advances, we find that existing DPO-based alignment of diffusion models still faces a fundamental limitation: simply maximizing the preference margin between ``winner'' and ``loser'' outputs does \emph{not} necessarily translate to better absolute generation quality of the finetuned model.

In our empirical analysis, we find that standard Diffusion-DPO~\cite{diffusion-dpo} exhibits unstable training dynamics, and the model’s generative quality can deteriorate as training proceeds. As illustrated in the left part of Fig.~\ref{fig:insight_combined}, we find that both the winner’s and loser’s denoising losses tend to increase over time, even though the preference margin ($\mathcal{L}^{w} - \mathcal{L}^{l}$) becomes more negative in the intended direction. This indicates that the model is widening the relative preference gap by making the less-preferred outputs worse, rather than truly improving the preferred outputs. In other words, relative alignment comes at the expense of absolute quality. The lack of a safeguard on the winner’s loss in existing DPO objectives leads to unstable training and potential collapse, corroborating observations in prior work~\cite{dpop,calibrated-dpo,see-dpo} that overly aggressive preference optimization can harm generative performance. These findings motivate the need for a new approach to preference-based diffusion finetuning that can increase preference alignment while preserving or improving the quality of the preferred outputs.

To address this challenge, we propose Diffusion-SDPO\footnote{Throughout the text, ``Diffusion\text{-}SDPO'' is used as a conceptual umbrella for our method and its guiding principles. When referring to concrete instantiations, we write ``X+SDPO'' to denote the integration of SDPO with a specific base DPO variant $X$ (e.g., Diffusion-DPO, DSPO, DMPO), which clarifies the application setting and configuration.} – a Safeguarded Direct Preference Optimization method for diffusion models. The key idea in Diffusion-SDPO is to introduce a simple yet effective winner-preserving update rule that controls the influence of the loser sample’s gradient at each training step. In contrast to standard DPO~\cite{llm_dpo,diffusion-dpo} which updates the model by contrasting winner and loser equally, we derive an adaptive scaling factor for the loser’s gradient based on the geometry of the winner and loser gradients. Intuitively, our method downweights the loser branch’s contribution whenever its gradient is misaligned with the winner’s gradient. Grounded in a first-order analysis, the safeguard computes a closed-form $\lambda_{\text{safe}}$ from the inner product of the winner and loser gradients, guaranteeing that each step does not worsen the preferred output’s reconstruction loss. In practice, Diffusion-SDPO seamlessly modifies the DPO objective with this adaptive loser scaling (see Fig.~\ref{fig:insight_combined}, right), which expands the preference margin while strictly controlling the absolute error of preferred outputs. Notably, our approach is model-agnostic and can be applied on top of various diffusion alignment frameworks~\cite{diffusion-dpo,mapo,dmpo,dspo}, acting as a plug-in optimizer that stabilizes training. Our contributions can be summarized as follows:
\begin{itemize}
    \item We show that enlarging the winner–loser margin in diffusion preference optimization does not guarantee higher quality and can degrade preferred outputs, revealing a gap between relative alignment and absolute error control.
    \item Based on these analysis, we propose \emph{Diffusion-SDPO}, a winner-preserving training scheme that adaptively scales the loser gradient by its geometric alignment with the winner gradient to first order. Our method is simple to implement and adds negligible overhead.
    \item Extensive experiments on SD\,1.5~\cite{sd15}, SDXL~\cite{sdxl} (both UNets~\cite{unet}), and the industrial-scale Ovis-U1~\cite{ovis-u1} (DiT~\cite{dit}) show that our method is architecture-agnostic. It delivers consistent improvements in preference metrics while preserving or enhancing aesthetic quality, stabilizing training, and avoiding collapse. These benefits persist across text-to-image, image editing and unified generation setups.
\end{itemize}

\section{Related Work}

\paragraph{Diffusion Models for Text-to-Image and Unified Generation.}
Diffusion models have become a leading paradigm for image synthesis, offering strong quality and diversity~\cite{diffusion_survey}. Denoising diffusion with a variational objective~\cite{ddpm} and continuous-time score-based formulations with SDEs~\cite{gm_sde,diffusion_sde} underpin modern systems. Refinements such as EDM~\cite{edm} and rectified flow or flow matching~\cite{rectified_flow,flow_matching} clarify objectives and improve robustness. Guidance-based conditioning~\cite{cg,cfg} enhances controllability. For text-to-image generation, latent diffusion~\cite{sd15} enables efficient high-resolution synthesis and supports large systems like SD3~\cite{sd3} and FLUX~\cite{flux}. In parallel, unified generators handle text-to-image and image editing within a single model~\cite{ovis-u1}. Our method applies to both families and is architecture-agnostic, working with UNet~\cite{unet}-style and DiT~\cite{dit}-style backbones.

\paragraph{Preference Optimization for Diffusion Models.}
Direct Preference Optimization~\cite{llm_dpo,diffusion-dpo,nc-dpo} has been adapted to diffusion models to align generation with human comparisons while avoiding full reinforcement learning. A broad class of variants~\cite{lpo,mapo} calibrates the preference margin or the relative branch influence to improve stability and protect the generation. Other approaches seek to guide the update directions and step magnitudes in LLMs~\cite{llm_survey} by employing subspace projections and modest objective clipping~\cite{orthogonal-dpo,bounded-dpo,calibrated-dpo,robust-dpo,chi2-po}. Related work such as DPOP~\cite{dpop} promotes positivity constraints to mitigate failure modes in preference optimization, and MaPPO~\cite{mappo} incorporates prior knowledge via a maximum-a-posteriori objective. Diffusion-specific methods further account for the multi-step nature of denoising by reweighting across timesteps or by adding entropy regularization, exemplified by Balanced-DPO~\cite{balanceddpo}, DSPO~\cite{dspo}, and SEE-DPO~\cite{see-dpo}. In contrast, our Diffusion-SDPO introduces a per-step, geometry-aware safe scaling factor based on the inner product between winner and loser output-space gradients, which provides direct control over the winner loss at each step while continuing to expand the preference margin.

\section{Preliminaries}
\paragraph{Diffusion Models.}
\label{sec:diffusion_models}
Diffusion models~\cite{diffusion,ddpm} construct a Markov chain that gradually corrupts clean data with additive noise and then learn a parametric denoiser to invert this corruption. Let a variance schedule $\{\beta_t\}_{t=1}^{T}$ be given and define $\alpha_t = 1-\beta_t$ and $\bar{\alpha}_t = \prod_{s=1}^{t}\alpha_s$. The forward process can be defined as:
\begin{equation}
\label{eq:forward_kernel}
q(x_t \mid x_{t-1}) \;=\; \mathcal{N}\!\bigl(x_t;\, \sqrt{\alpha_t}\,x_{t-1},\, (1-\alpha_t)\,\mathbf{I}\bigr),
\end{equation}
which implies the following closed-form perturbation of a clean sample $x_0$:
\begin{equation}
\label{eq:closed_form}
x_t \;=\; \sqrt{\bar{\alpha}_t}\,x_0 \;+\; \sqrt{1-\bar{\alpha}_t}\,\epsilon, 
\qquad \epsilon \sim \mathcal{N}(\mathbf{0},\mathbf{I}).
\end{equation}
Equivalently, the marginal distribution conditioned on $x_0$ is
\begin{equation}
\label{eq:marginal}
q(x_t \mid x_0) \;=\; \mathcal{N}\!\bigl(x_t;\, \sqrt{\bar{\alpha}_t}\,x_0,\, (1-\bar{\alpha}_t)\,\mathbf{I}\bigr).
\end{equation}

Learning proceeds by training a network $\epsilon_{\theta}$ that receives the noised input $x_t$ and the time index $t$ to predict the injected noise. Using the reparameterization in Eq.~\ref{eq:closed_form}, the standard objective minimizes mean squared error between the true noise and the prediction:
\begin{equation}
\label{eq:noise_mse}
\mathcal{L}_{\text{diffusion}} 
\;=\; \mathbb{E}_{x_0, t, \epsilon}
\;\bigl\|\, \epsilon_{\theta}\!\bigl(x_t, t\bigr) - \epsilon \bigr\|_2^2.
\end{equation}
where $x_0 \!\sim\! p_{\text{data}}, t \!\sim\! \mathrm{Uniform}\{1,\dots,T\}$ and $\epsilon \!\sim\! \mathcal{N}(\mathbf{0},\mathbf{I})$. Minimizing $\mathcal{L}_{\text{diffusion}}$ yields a time-aware denoiser that can be applied in reverse order to iteratively remove noise and synthesize new samples from an initial Gaussian latent.

\paragraph{Diffusion Model Alignment via Preference.}
Given a prompt $c$ and two images $x^w_0$ (preferred, ``winner'') and $x^l_0$ (less preferred, ``loser''), preference alignment for diffusion models seeks parameters $\theta$ such that the model assigns higher likelihood to $x^{w}_0$ than to $x^{l}_0$ \cite{nc-dpo,diffusion-dpo}. A diffusion sampler produces a trajectory $(x_T,\ldots,x_0)$ and, at each time $t$, a reverse conditional $p_\theta(x_t \mid x_{t+1},c)$~\cite{ddpm,diffusion,rectified_flow}. To instantiate DPO in this setting, we adopt the standard formulation wherein the stepwise preference score is the log-likelihood ratio with respect to a frozen reference model~\cite{diffusion-dpo,dspo}:
\begin{equation}
\label{eq:step-reward}
r_t(x_t,c)\;=\;\beta\,\log\frac{p_\theta(x_t \mid x_{t+1},c)}{p_{\text{ref}}(x_t \mid x_{t+1},c)}\,.
\end{equation}
The Diffusion–DPO~\cite{diffusion-dpo} loss applies Bradley-Terry-style~\cite{BT_model} logistic regression to the winner-loser pair at the same $t$:
\begin{equation}
\label{eq:diffusion-dpo-step}
\mathcal{L}_{\text{Diffusion-DPO}}
\;=\;
-\;\mathbb{E}\Big[
\log \sigma\!\Big(
r_t(x_t^{w},c)\;-\;r_t(x_t^{l},c)
\Big)\Big],
\end{equation}
and averages Eq.~\ref{eq:diffusion-dpo-step} over $t\!\in\!\{0,\ldots,T\!-\!1\}$ (or samples a single $t$ per pair for an unbiased stochastic estimator). Equivalently, substituting Eq.~\ref{eq:step-reward} into Eq.~\ref{eq:diffusion-dpo-step} gives the explicit form
\begin{equation}
\label{eq:diffusion-dpo-expanded}
\begin{split}
\mathcal{L}_{\text{Diffusion-DPO}}
= &-\,\mathbb{E}\!\Bigl[\log \sigma\Bigl(
\beta \log \frac{p_\theta(x_t^{w}\mid x_{t+1}^{w},c)}{p_{\text{ref}}(x_t^{w}\mid x_{t+1}^{w},c)}
\\
\qquad &- \beta \log \frac{p_\theta(x_t^{l}\mid x_{t+1}^{l},c)}{p_{\text{ref}}(x_t^{l}\mid x_{t+1}^{l},c)}
\Bigr)\Bigr].
\end{split}
\end{equation}
Under common parameterizations, Eq.~\ref{eq:step-reward} reduces to simple residual comparisons. For DDPM-style Gaussians \cite{ddpm,diffusion}, writing $\hat\epsilon_\theta = \epsilon_\theta(x_{t+1},c,t)$ for the predicted noise, $\hat\epsilon_\text{ref} = \epsilon_\text{ref} (x_{t+1},c,t)$ for the reference noise and $\epsilon$ for the ground-truth noise that forms $x_t$, the log-ratio can be expressed as:
\begin{equation}
\label{eq:gaussian-logratio}
\log\!\frac{p_\theta(x_t \mid x_{t+1},c)}{p_{\text{ref}}(x_t \mid x_{t+1},c)}
\!\!\propto\!\!
-\tfrac{1}{2}\!\big\|\hat\epsilon_\theta-\epsilon\big\|_2^2
\!\!+\!\tfrac{1}{2}\!\big\|\hat\epsilon_{\text{ref}}-\epsilon\big\|_2^2
\!+\!\text{const},
\end{equation}
and an analogous expression holds for velocity or flow-matching parameterizations by replacing the noise residual with the corresponding target~\cite{rectified_flow}. For notational brevity, we write the stepwise contrastive objective as $\mathcal{L}(x_{t+1},c,t)=\frac{1}{2}\|\epsilon_\theta(x_{t+1},c,t) - \epsilon\|_2^2
-\frac{1}{2}\|\epsilon_{\text{ref}}(x_{t+1},c,t) - \epsilon\|_2^2$. Hence, the winner and loser margin loss are defined as $\mathcal L^w=\mathcal{L}(x_{t+1}^{w},c,t)$ and $\mathcal L^l=\mathcal{L}(x_{t+1}^{l},c,t)$, respectively. Substituting Eq.~\ref{eq:gaussian-logratio} into Eq.~\ref{eq:diffusion-dpo-expanded} gives the training loss
\begin{equation}
\label{eq:instantiate_dpo}
 \mathcal{\hat L}_{\text{Diffusion-DPO}}\!
= \!-\!\,\mathbb{E}_{t,\epsilon,c,x_0^w, x_0^l}\!\left[\log \sigma\!\Big(\!\!-\!\beta\!\,(\mathcal{L}^{w}-\mathcal{L}^{l})\Big)\right]\!,
\end{equation}
where $\mathcal D=\{(c,x_{0}^{w},x_{0}^{l})\}$ denotes the DPO training dataset.

\paragraph{Limitations of Standard DPO.}
Substituting Eq.~\ref{eq:gaussian-logratio} into Eq.~\ref{eq:diffusion-dpo-expanded} yields an implementable objective whose inner term is the per-step error difference between winner and loser branches. Diffusion–DPO~\cite{diffusion-dpo} thus encourages decreasing the winner’s prediction error while increasing the loser’s at the same timestep. However, this objective does not guarantee a monotonic decrease of the winner loss. Empirically, over-penalizing the loser can also worsen the preferred sample. In the left part of Fig.~\ref{fig:insight_combined}, the margin $\mathcal{L}^{w}-\mathcal{L}^{l}$ becomes increasingly negative, yet both $\mathcal{L}^{w}$ and $\mathcal{L}^{l}$ increase, indicating degradation of absolute performance and potential instability or collapse. This exposes a gap between \emph{relative} alignment (widening the margin) and \emph{absolute} error control (preserving the preferred sample). The difficulty is that the winner and loser gradients are misaligned and vary across timesteps. We therefore introduce a simple stepwise update that, to first order, guarantees the preferred loss does not increase at each step while still promoting margin expansion.

\section{Method: Diffusion-SDPO (Safe DPO)}

We propose \textbf{Diffusion-SDPO}, a novel preference optimization scheme that adds a safety guard to the DPO update. The method adaptively scales the influence of the loser branch by a time-dependent factor $\lambda_t$ so that the preferred sample’s loss $\mathcal L^w$ does not increase after each parameter update. In practice, we follow the standard Diffusion–DPO pipeline: given a prompt $c$ and a pair $(x_0^w, x_0^l)$, we compute the per-sample losses
$\mathcal L^w$ and $\mathcal L^l$ at the same diffusion time $t$, and then modify the backpropagated update by multiplying the loser-branch gradient by the safety factor to enforce a safe update condition. This directly addresses the limitation discussed above, because preventing any increase in the preferred loss ensures that preference-driven updates do not degrade the preferred output while still improving the preference margin.

\subsection{Safe Update via First-Order Approximation}
Our objective is to ensure that a gradient update driven by the preference loss (cf. Eq.~\ref{eq:diffusion-dpo-expanded}) does not increase the winner’s loss. For clarity of exposition, consider a linearized preference objective combining the two branches:\footnote{In practice, the actual Diffusion-DPO gradient (Eq.~\ref{eq:diffusion-dpo-step}) includes a logistic scaling factor $\sigma(\cdot)$ that multiplies the winner and loser gradients equally, thus not altering the update direction. We therefore analyze the simpler weighted difference objective that captures the same first-order direction.}
\begin{equation}
\mathcal{L}^{\text{pref}}(\theta) = \mathcal L^w(\theta)-\lambda \cdot\mathcal L^l(\theta),
\label{eq:pref-obj-linear}
\end{equation}
where $\lambda>0$ is a scalar that adjusts the relative weight on the loser’s loss. Setting $\lambda=1$ recovers the intuitive gradient direction of standard DPO (decrease $\mathcal L^w$, increase $\mathcal L^l$), while $\lambda>1$ would place even more emphasis on penalizing the loser. Our goal is to find an upper bound on $\lambda$ that guarantees $\mathcal L^w$ will not increase for an infinitesimal gradient step on $\mathcal{L}^{\text{pref}}$.

Let $\nabla_\theta \mathcal L^w$ and $\nabla_\theta \mathcal L^l$ denote the gradients of the winner and loser losses, respectively. A gradient descent step of size $\eta$ on Eq.~\ref{eq:pref-obj-linear} gives the parameter update:
\begin{equation}
\Delta \theta = -\eta\cdot\nabla_\theta \mathcal{L}^{\text{pref}} = -\eta\Big(\nabla_\theta \mathcal L^w - \lambda\nabla_\theta \mathcal L^l\Big).
\end{equation}
The first-order change in the winner’s loss can be approximated by a Taylor expansion:
\begin{equation}
\Delta \mathcal L^w \!\!\approx\!\! \nabla_\theta {\mathcal L^w}^\top\!\!\Delta \theta \!=\! \!-\eta\Big(\!\|\nabla_\theta \mathcal L^w\|_2^2 \!-\! \lambda\nabla_\theta {\mathcal L^w}^\top\!\nabla_\theta \mathcal L^l\!\Big).
\label{eq:delta-Lw}
\end{equation}
To \emph{prevent} increase in $\mathcal L^w$, we require $\Delta \mathcal L^w \le 0$, i.e., $\nabla_\theta {\mathcal L^w}^\top \Delta \theta \le 0$. Ignoring the trivial positive factor $\eta$, the safety condition becomes:
\begin{equation}
\|\nabla_\theta \mathcal L^w\|_2^2 - \lambda\nabla_\theta {\mathcal L^w}^\top\nabla_\theta \mathcal L^l \ge 0.
\label{eq:safety-cond}
\end{equation}
Solving for $\lambda$ yields a bound on the allowable loser weight:
\begin{equation}
\lambda \le \frac{\|\nabla_\theta \mathcal L^w\|_2^2}{\nabla_\theta {\mathcal L^w}^\top\nabla_\theta \mathcal L^l}.
\label{eq:lambda-bound-param}
\end{equation}
Notably, if the dot product $\nabla_\theta {\mathcal L^w}^\top \nabla_\theta \mathcal L^l$ is negative or zero, then Eq.~\ref{eq:safety-cond} is automatically satisfied for any $\lambda \ge 0$. In those cases, the update is intrinsically safe: the loser branch either helps reduce $\mathcal L^w$ or affects orthogonal parameter directions. The problematic scenario is when $\nabla_\theta {\mathcal L^w}^\top \nabla_\theta \mathcal L^l > 0$, i.e., the loser’s gradient has a component that would raise the winner’s loss. Eq.~\ref{eq:lambda-bound-param} then yields a finite positive $\lambda$ threshold. Any choice of $\lambda$ above this threshold would violate the safety inequality, leading to $\Delta \mathcal L^w > 0$ to first order. Conversely, choosing $\lambda$ at or below this threshold ensures $\Delta \mathcal L^w \approx 0$ or negative, guaranteeing that the winner’s loss does not increase.

\begin{algorithm}[t]
\caption{Training of Diffusion-SDPO.}
\label{alg:sdpo}
\KwIn{Dataset $\mathcal D=\{(c,x_{0}^{w},x_{0}^{l})\}$; model $\epsilon_\theta$; reference $\epsilon_{\text{ref}}$; safety slack $\mu\!\in\![0,1]$; schedule length $T$; learning rate $\eta$.}

\While{not converged}{
  1. Sample $t \sim \mathrm{Uniform} \{0,\dots,T-1\} ,\epsilon \sim \mathcal{N}(\mathbf{0},\mathbf{I}), (c,x_{0}^w,x_{0}^l) \sim \mathcal D$.
  
  2. Get $(x_{t+1}^w, x_{t+1}^l)$ from Eq.~\ref{eq:closed_form} and compute
  \vspace{-6pt}
  \begin{align*}
    \hat\epsilon^{w}_\theta&=\epsilon_\theta(x_{t+1}^{w},c,t), &&\hat\epsilon^{l}_\theta\,\;=\epsilon_\theta(x_{t+1}^{l},c,t), \\ \hat\epsilon^{w}_\text{ref}&=\epsilon_\text{ref}(x_{t+1}^{w},c,t),
    &&\hat\epsilon^{l}_\text{ref}=\epsilon_\text{ref}(x_{t+1}^{l},c,t).
  \end{align*}
  \vspace{-18pt}
  
  3. Get per-branch residual objectives:
  \vspace{-6pt}
  \begin{align*}
  &\mathcal{L}^{w} =\tfrac12\|\hat\epsilon^{w}_\theta -\epsilon\|_2^2-\tfrac12\|\hat\epsilon^{w}_{\text{ref}}-\epsilon\|_2^2, \\
  &\mathcal{L}^{l} \;=\tfrac12\|\hat\epsilon^{l}_\theta -\epsilon\|_2^2-\tfrac12\|\hat\epsilon^{l}_{\text{ref}}-\epsilon\|_2^2.
  \end{align*}
  \vspace{-18pt}

  \textcolor{red}{4. Compute $\lambda_\text{safe}$ using Eq.~\ref{eq:true_lambda_safe}: $\displaystyle \lambda_{\text{safe}} = (1-\mu)\|g^w\|_2^2 / {g^w}^\top g^l$.}

  \textcolor{red}{5. Scale only loser gradients: $\;\mathcal{L}^{l}_{\text{scaled}}
=\mathcal{L}^{l}_{\text{detach}}+\lambda_\text{safe}\big(\mathcal{L}^{l}-\mathcal{L}^{l}_{\text{detach}}\big)^\dagger$.}


  6. Build loss using Eq.~\ref{eq:instantiate_dpo}: $\mathcal{L}_{\text{DPO}} = -\log \sigma \left( -\beta (\mathcal L^w - \mathcal L^l_\text{scaled}) \right)$.

  7. Update $\theta \leftarrow \theta - \eta \,\nabla_{\theta}\, \mathcal{L}_{\text{DPO}}$.
}
\KwOut{Finetuned model $\epsilon_\theta$.}
{\footnotesize $\dagger:\mathcal{L}^{l}_{\text{detach}}$ is a copy of $\mathcal{L}^{l}$ without gradient flow.}
\end{algorithm}

\subsection{Closed-Form Safeguard in Output Space}
Directly evaluating the parameter-space bound in Eq.~\ref{eq:lambda-bound-param} is infeasible for a high-dimensional model, since it would require computing and storing the full gradients $\nabla_\theta \mathcal L^w$ and $\nabla_\theta \mathcal L^l$ just to take their dot product. However, we can derive a convenient proxy by considering gradients in the model’s \emph{output space}. Modern diffusion models predict a noise or image tensor as output, and the training loss (e.g., a denoising score-matching loss~\cite{ddpm}) is defined on this output. Let $o^w$ and $o^l$ denote the model’s output activations for the winner and loser branches respectively (for example, $o$ could be the predicted noise residual at a certain diffusion step). Using the chain rule, we have $\nabla_\theta \mathcal L^w = {J^w}^\top \nabla_{o} \mathcal L^w$ and $\nabla_\theta \mathcal L^l = {J^l}^\top \nabla_{o} \mathcal L^l$, where $J$ is the Jacobian $\partial o/\partial \theta$ and $\nabla_{o} \mathcal L$ is the gradient of the loss with respect to the model output. Let $g^w = \nabla_{o}\mathcal{L}^{w}$ and $g^l = \nabla_{o}\mathcal{L}^{l}$ denote the output-space gradients for the winner and the loser. Eq.~\ref{eq:lambda-bound-param} can then be written as:
\begin{align}
\lambda &\le \!\frac{\|\nabla_\theta \mathcal L^w\|_2^2}{\nabla_\theta {\mathcal L^w}^\top\nabla_\theta \mathcal L^l}
= \frac{{g^w}^\top \!\big({J^w} {J^w}^\top\big) {g^w}}{{g^w}^\top \!\big({J^w} {J^l}^\top\big) {g^l}} \\
&=\! \frac{\|{g^w}\|_2^2}{\,{g^w}^\top {g^l}\,} \!\cdot\!
   \underbrace{
   \frac{\,{g^w}^\top \!\big({J^w} {J^w}^\top\big) {g^w}\,}{\,\|{g^w}\|_2^2\,}
   \!\!\bigg/\!\!
   \frac{\,{g^w}^\top \!\big({J^w} {J^l}^\top\big) {g^l}\,}{\,{g^w}^\top {g^l}\,}
   }_{\rho}\!.
\label{eq:grad-dot-jacobian}
\end{align}

The factor $\rho$ in Eq.~\ref{eq:grad-dot-jacobian} encodes local Jacobian geometry. Estimating $\rho$ during training requires parameter–space backpropagation and adds memory usage or wall–clock time (cf. Table~\ref{tab:unified_wpr}). To keep the update lightweight, we do not track $\rho$ explicitly. Instead, we absorb it into a scalar \emph{safety slack} $\mu\in[0,1]$ that contracts the proxy in a controlled way:
\begin{equation}
\label{eq:true_lambda_safe}
\lambda_{\text{safe}}=\frac{(1-\mu)\,\|g^w\|_2^2}{{g^w}^\top g^l}.
\end{equation}
This removes any dependence on parameter–space Jacobians and uses only the output–space gradients ${g^w}$ and ${g^l}$ that are already computed for the loss, so the additional cost is negligible. The slack $\mu$ serves as a robust guardrail under arbitrary local geometry. Larger $\mu$ yields a more conservative scaling of the loser branch; smaller $\mu$ recovers a more aggressive update. In practice, we find a fixed $\mu$ works well (see Fig.~\ref{fig:param_analysis} for ablation on $\mu$), and for an appropriate choice of $\mu$, the output-space scheme yields $\lambda_{\text{safe}}$ trajectories that closely match those obtained from parameter-space gradients (cf.~Fig.~\ref{fig:lambda_ablation}).

During training, we clip $\lambda_\text{safe}$ to $[0,1]$ for stability (if ${g^w}^\top g^l\le 0$, we set $\lambda_{\text{safe}}=1$). Whenever the loser’s error vector has a positive correlation with the winner’s error vector (${g^w}^\top g^l > 0$), $\lambda_{\text{safe}}$ provides a finite limit to how strongly we can apply the loser’s gradient without risking an increase in the winner’s loss. For the logistic DPO objective, we implement this by scaling the backpropagated loser gradient with $\lambda_{\text{safe}}$. Algorithm~\ref{alg:sdpo} summarizes the procedure to integrate SDPO into Diffusion-DPO. For other methods, $\lambda_\text{safe}$ is similarly used to scale the loser branch.

\begin{table}[t]
\centering
\small
\setlength{\tabcolsep}{0.15pt}
\caption{Reward score comparison on the HPS V2 with SD 1.5. Rows labeled ``+ SDPO'' report the performance obtained by applying our SDPO to the corresponding base method in the preceding row. $^\dagger$: results from our implementation due to the lack of official code. Best results are in \textbf{bold}. Owing to space constraints, the full table is provided in Table~\ref{tab:score_15_full}.}
\begin{tabular}{l|ccccc}
    \hline
    \textbf{Method} 
    & \textbf{PickScore($\uparrow$)} 
    & \textbf{\hphantom{0} HPS($\uparrow$) \hphantom{0}} 
    & \textbf{Aes.($\uparrow$)} 
    & \textbf{\hphantom{0} CLIP($\uparrow$) \hphantom{0}} 
    & \textbf{IR($\uparrow$)} \\
    \hline
    SD 1.5 & 0.2088 & 0.2697 & 5.4933 & 0.3480 & -0.0469 \\
    SFT & 0.2168 & 0.2838 & 5.7851 & 0.3591 & 0.6619 \\
    Diff.-KTO & 0.2164 & 0.2766 & 5.6288 & 0.3420 & 0.5593 \\
    MaPO$^\dagger$ & 0.2124 & 0.2760 & 5.6890 & 0.3528 & 0.3308 \\
    DPOP$^\dagger$ & 0.2144 & 0.2780 & 5.7071 & 0.3563 & 0.3735 \\ 
    \hdashline
    Diff.-DPO & 0.2131 & 0.2743 & 5.6639 & 0.3552 & 0.1705 \\
    \hphantom{0} + \textbf{SDPO}  & 0.2174 & 0.2827 & \bf 5.8744 & 0.3600 & 0.6211  \\
    \hdashline
    DSPO & 0.2168 & 0.2837 & 5.8346 & 0.3598 & 0.6483 \\
    \hphantom{0} + \textbf{SDPO} & 0.2172 & 0.2847 & 5.8474 & 0.3586 & 0.6578\\
    \hdashline
    DMPO$^\dagger$ & 0.2131 & 0.2766 & 5.6538 & 0.3551 & 0.3171 \\
    \hphantom{0} + \textbf{SDPO} & \textbf{0.2182} & \textbf{0.2848} & 5.8574 & \textbf{0.3612} & \textbf{0.7061} \\
    \hline
\end{tabular}
\label{tab:score_15}
\end{table}

\begin{table}[t]
\centering
\small
\setlength{\tabcolsep}{0.15pt}
\caption{Reward score comparison on the HPS V2 with SDXL. $^\dagger$: results from our implementation due to the lack of official code. The full table is provided in Table~\ref{tab:score_xl_full}.}
\begin{tabular}{l|ccccc}
    \hline
    \textbf{Method} 
    & \textbf{PickScore($\uparrow$)} 
    & \textbf{\hphantom{0} HPS($\uparrow$) \hphantom{0}} 
    & \textbf{Aes.($\uparrow$)} 
    & \textbf{\hphantom{0} CLIP($\uparrow$) \hphantom{0}} 
    & \textbf{IR($\uparrow$)} \\
    \hline
    SDXL & 0.2290 & 0.2900 & 6.1271 & 0.3847 & 0.9047 \\
    SFT  & 0.2228 & 0.2883 & 5.9689 & 0.3806 & 0.8528 \\
    MaPO & 0.2293 & 0.2934 & \textbf{6.1882} & 0.3840 & 0.9703 \\
    \hdashline
    Diff.-DPO & 0.2288 & 0.2927 & 6.1380 & 0.3840 & 1.0159 \\
    \hphantom{0} + \textbf{SDPO} & 0.2308 & 0.2938 & 6.1284 & 0.3879 & 1.0326 \\
    \hdashline
    DSPO & 0.2273 & 0.2916 & 6.0424 & 0.3894 & 1.0054 \\
    \hphantom{0} + \textbf{SDPO} & 0.2293 & \textbf{0.2944} & 6.1040 & 0.3889 & \textbf{1.0745} \\
    \hdashline
    DMPO$^\dagger$ & 0.2302 & 0.2921 & 6.1101 & 0.3875 & 1.0154 \\
    \hphantom{0} + \textbf{SDPO} & \textbf{0.2308} & 0.2933 & 6.1113 & \textbf{0.3897} & 1.0521 \\
    \hline
\end{tabular}
\label{tab:score_xl}
\end{table}

\begin{table*}
\centering
\small
\setlength{\tabcolsep}{5pt}
\caption{Average win rate comparison (\%) over the HPS V2 using SD 1.5. Each row reports \emph{Model 1} vs.\ \emph{Model 2} on identical prompts. The upper block summarizes SDPO augmentation results (base + SDPO vs.\ base), and the lower block compares each model against SD\,1.5. Values $>50\%$ indicate that \emph{Model 1} generally outperforms \emph{Model 2}.}

\begin{tabular}{l|l|cccccc}
\hline
\textbf{Model 1} & \textbf{Model 2} &
\textbf{PickScore} & \textbf{HPS V2} & \textbf{Aes.} &  \textbf{CLIP} & \textbf{ImageReward} & \textbf{Mean} \\
\hline
\multicolumn{8}{l}{\textit{SDPO augmentation effect (base+SDPO vs base)}}\\
\hline
\makebox[4em][l]{Diff.-DPO} + SDPO    & Diff.-DPO & 66.12 & 78.62 & \bf 70.62 & 52.50 & \bf 71.62 & 67.90 \\
\makebox[4em][l]{DSPO}     + SDPO    & DSPO & 52.62 & 53.00 & 53.25 & 48.50  & 52.38 & 51.95 \\
\makebox[4em][l]{DMPO}     + SDPO    & DMPO & \bf 73.50 & \bf 79.25 &  67.12 & \bf 53.12  & 71.50 & \bf 68.90 \\
\hline
\hline
\multicolumn{8}{l}{\textit{Versus SD 1.5}}\\
\hline
Diff.-DPO        & SD 1.5  & 76.38 & 69.50 & 66.88 & 57.50  & 63.50 & 66.75 \\
\makebox[4em][l]{Diff.-DPO} + SDPO & SD 1.5  & 80.75 & 86.25 & \bf 83.38 & 56.88  & 79.25 & 77.30 \\
DSPO & SD 1.5 &78.38 & 86.00 & 78.75 & \bf 58.88 & 79.75 & 76.35 \\
\makebox[4em][l]{DSPO} + SDPO     & SD 1.5 & 81.75 & \bf 89.75 & 79.12  & 56.75 & 80.12 & 77.50 \\
DMPO            & SD 1.5  & 68.50 & 74.50 & 68.38 & 53.00  & 69.88 & 66.85 \\
\makebox[4em][l]{DMPO} + SDPO     & SD 1.5  & \bf 82.25 & 88.12 & 79.25 & 58.38  & \bf 81.75 & \bf 77.95 \\
\hline
\end{tabular}
\label{tab:hpsv2_winrate_sd15}
\end{table*}

\section{Experiments}
\subsection{Experimental Setting}
\label{sec:emperimental_setting}

\paragraph{Datasets and Models.}
Following~\cite{dspo,dmpo}, we finetune Stable Diffusion~1.5 (SD 1.5) and SDXL on preference pairs from Pick-a-Pic V2 (Pick V2)~\cite{pap} training set. For evaluation, we use the test prompts from Pick V2, HPS V2~\cite{hpsv2}, and PartiPrompts~\cite{partiprompts}. Beyond SD 1.5 and SDXL, we also conduct experiments on Ovis-U1~\cite{ovis-u1} (3.6B), a DiT~\cite{dit} model trained in a unified manner to support both text-to-image synthesis and image editing. To enable DPO finetuning on Ovis-U1, we construct a mixed preference corpus that integrates text-to-image and editing pairs, totaling about 33K pairs.

\paragraph{Training Details and Baselines.}
We integrate SDPO into Diffusion-DPO~\cite{diffusion-dpo}, DSPO~\cite{dspo}, and DMPO~\cite{dmpo} implementations and keep their official hyperparameters. All models are finetuned for 2000 steps with a global batch size of 2048. The learning rate is $1\times10^{-8}$ for SD 1.5 and $1\times10^{-9}$ for SDXL. For the safeguard coefficient $\mu$, on SD 1.5 we set $0.9$ for Diffusion\text{-}DPO+SDPO and DMPO+SDPO, $0.2$ for DSPO+SDPO. On SDXL, $\mu$ is fixed as $0.6$ for all variants. We compare against several baselines: the original pretrained SD 1.5 and SDXL, supervised finetuning (SFT), Diffusion-KTO~\cite{diffusion-kto}, MaPO~\cite{mapo}, DPOP~\cite{dpop}, and original Diffusion-DPO, DSPO, DMPO. For baselines we follow a strict hierarchy. If official checkpoints are publicly available, we evaluate those directly. If checkpoints are unavailable but official code exists, we run the released implementation with the authors’ recommended settings. If neither is available, we reimplement the method from the paper.

\paragraph{Evaluation.}
We evaluate models on automatic preference metrics, including PickScore~\cite{pap}, HPS V2~\cite{hpsv2}, LAION Aesthetic Classifier~\cite{laion-aesthetics}, CLIP~\cite{clip} and ImageReward~\cite{imagereward} scores. Sampling uses a guidance scale of 7.5 and 50 denoising steps. For Ovis-U1, we additionally evaluate structured text-to-image alignment on GenEval~\cite{geneval} and DPG-Bench~\cite{dpg_bench}, as well as image-editing performance on ImgEdit~\cite{imgedit} and GEdit-EN~\cite{gedit}.

\subsection{Main Results}

Table~\ref{tab:score_15},\ref{tab:score_xl} show that adding SDPO to Diffusion\mbox{-}DPO, DSPO, and DMPO consistently improves automatic reward metrics under SD~1.5 and SDXL, with DMPO+SDPO typically giving the best overall scores. Win\mbox{-}rate results on SD~1.5 (Table~\ref{tab:hpsv2_winrate_sd15}) further confirm that each base method benefits from SDPO and that SDPO variants also outperform the SD~1.5 baseline, indicating stronger preference alignment without loss of quality. On SDXL, the gains are moderate yet consistent (see Table~\ref{tab:hpsv2_winrate_sdxl}), suggesting reliable scaling to larger UNet~\cite{unet} backbones. 

\begin{figure}
  \centering
  \includegraphics[width=.99\linewidth]{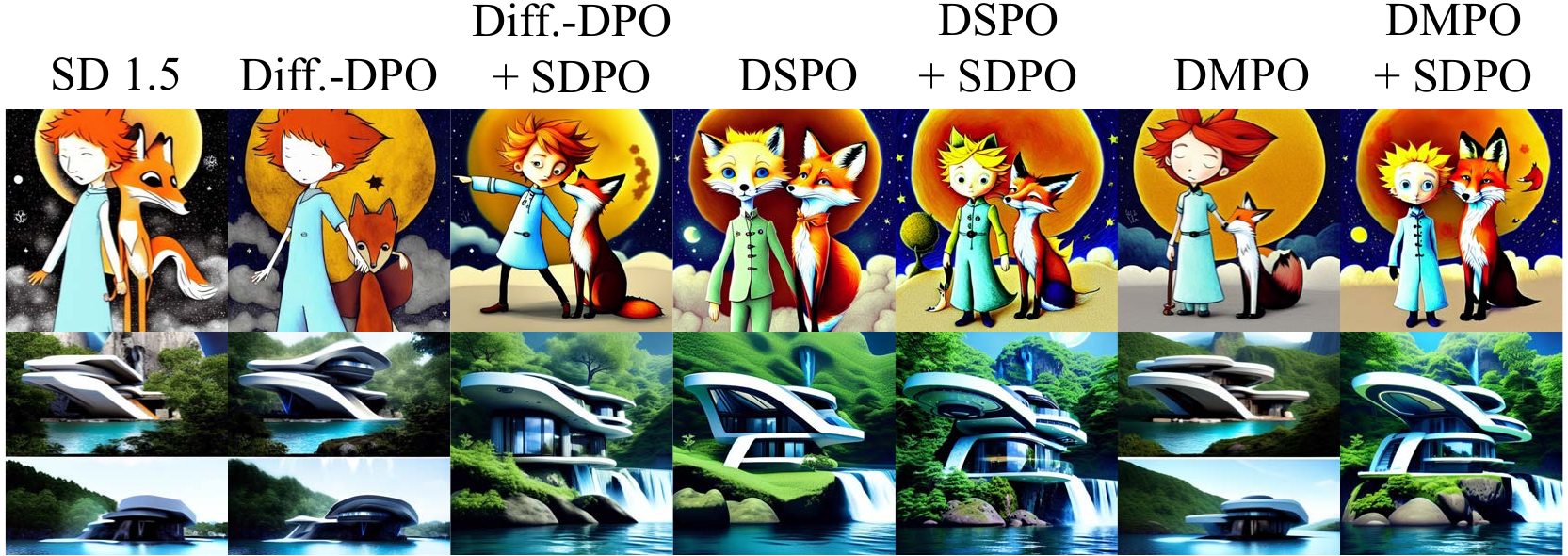}
\caption{Qualitative comparison of different methods using SD 1.5. Prompt: \textit{1) The Little Prince and the fox in a Tim Burton style artwork. 2) A futuristic modern house on a floating rock island surrounded by waterfalls, moons, and stars on an alien planet.} See Fig.~\ref{fig:images_sd15_appendix} for more results.}
  \label{fig:images_sd15}
\end{figure}

On the unified Ovis\mbox{-}U1 model (Table~\ref{table:ovis_u1}), SDPO yields clear improvements in preference metrics and editing scores, demonstrating effectiveness on a DiT backbone as well. While naive Diffusion-DPO can enlarge preference margins at the expense of fidelity, our safeguarded integrations preserve details and improve prompt adherence across diverse prompts. The visual evidence (cf. Fig.~\ref{fig:images_sd15},~\ref{fig:images_sd15_appendix},~\ref{fig:images_ovis_u1}) aligns with the quantitative trends, indicating that SDPO stabilizes optimization and enhances perceptual quality.

\begin{figure}
  \centering
  \includegraphics[width=.99\linewidth]{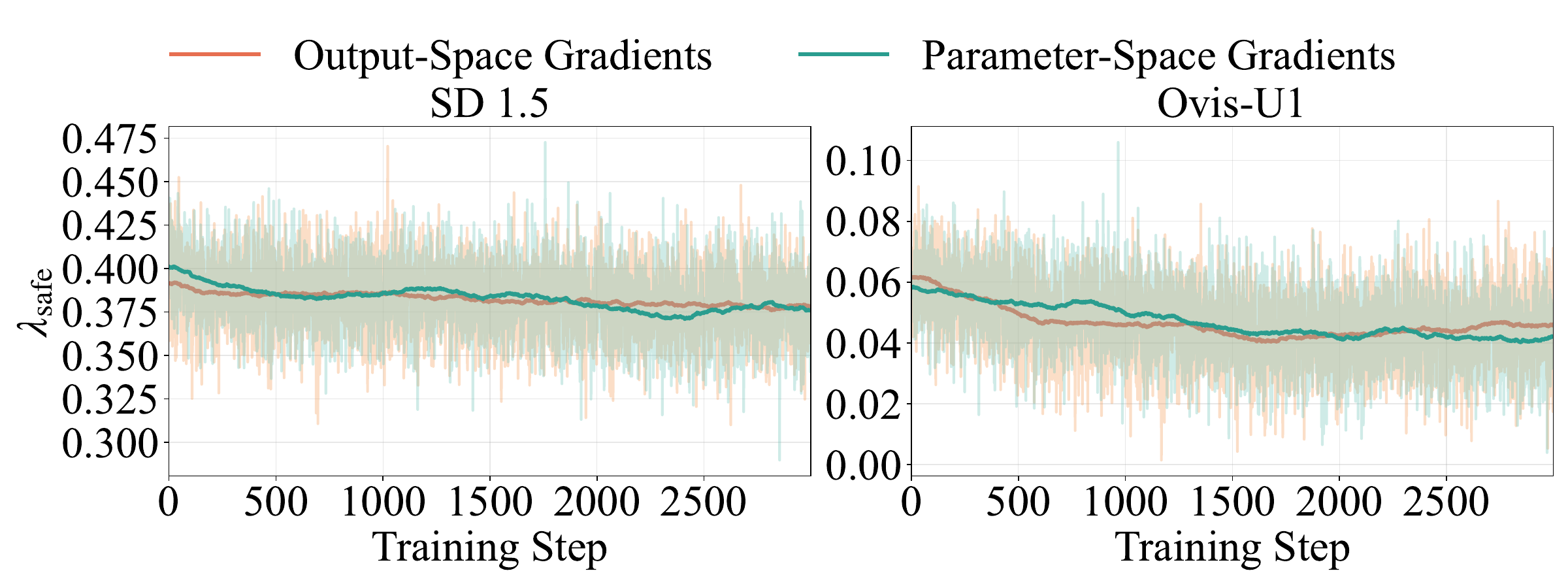}
   \caption{Training dynamics of $\lambda_{\text{safe}}$ on SD 1.5 (left) and Ovis-U1 (right) with two computation schemes (using output-space gradients \textit{vs.} parameter-space gradients). The trajectories closely match throughout training, and the output-space variant requires substantially less computation while maintaining comparable aesthetic rewards (see Table~\ref{tab:unified_wpr}).}
  \label{fig:lambda_ablation}
\end{figure}

\begin{table}
\small
\centering
\caption{Ablation results for winner-preserving rules on SD\,1.5 (prompts: HPS\,V2). We report PickScore/HPS\,V2, one-step training Time (s) and peak GPU memory (GB) on a single NVIDIA A100 with batch size 16 and 128 gradient accumulation steps in BF16. $^\ddagger$: fixed $\lambda_{\text{safe}}$ in SDPO. $^\dagger$: $\lambda_{\text{safe}}$ computed with parameter-space gradients.}
\label{tab:unified_wpr}
\setlength{\tabcolsep}{0.5pt}
\renewcommand{\arraystretch}{0.95}
\begin{tabular}{l|cc|cc}
\hline
\textbf{Method} & \textbf{PickS. ($\uparrow$)} & \textbf{HPS ($\uparrow$)} & \textbf{Time ($\downarrow$)} & \textbf{GPU Mem. ($\downarrow$)} \\
\hline
MaPO                               & 0.2124 & 0.2760 & 162    & 53.8  \\
DPOP                               & 0.2144 & 0.2780 & 187    & 58.5  \\
Diff.-DPO                          & 0.2131 & 0.2743 & 183    & 57.0  \\
Diff.-DPO+SDPO$^\ddagger$          & 0.2158 & 0.2803 & 183    & 57.0  \\
Diff.-DPO+SDPO$^\dagger$           & 0.2176 & 0.2828 & 314    & 63.3  \\
Diff.-DPO+SDPO                     & 0.2174 & 0.2827 & 184    & 57.1  \\
\hline
\end{tabular}
\end{table}

\begin{table*}
\centering
\small
\caption{Comparison of Ovis-U1~\cite{ovis-u1} variants on preference, structured alignment, and image editing benchmarks. Higher is better ($\uparrow$). SDPO is particularly effective for preference alignment in large-scale models.}
\setlength{\tabcolsep}{1.5pt}
\begin{tabular}{l|cc|cc|cc}
\hline
\multirow{2}{*}{\textbf{Model}} 
& \multicolumn{2}{c|}{\textbf{Preference Eval} ($\uparrow$)} 
& \multicolumn{2}{c|}{\textbf{Structured Alignment Eval} ($\uparrow$)} 
& \multicolumn{2}{c}{\textbf{Image Editing} ($\uparrow$)} \\
\cline{2-7}
& CLIP & HPS V2 & GenEval & DPG-Bench & ImgEdit & GEdit-EN \\
\hline
Ovis-U1        & 0.3188 & 0.2986 & 0.89 & 83.72 & 4.00 & 6.42 \\
Ovis-U1 + DPO  & 0.3192 & 0.2997 & 0.88 & 83.78 & 4.01 & 6.43 \\
Ovis-U1 + SDPO & \bf 0.3201 & \bf 0.3082 & \bf 0.89 &  \bf 84.84 & \bf 4.11 & \bf 6.60 \\
\hline
\end{tabular}
\label{table:ovis_u1}
\end{table*}

\subsection{Ablation Study}
\label{sec:ablation_study}

\begin{figure*}
  \centering
  \includegraphics[width=.99\linewidth]{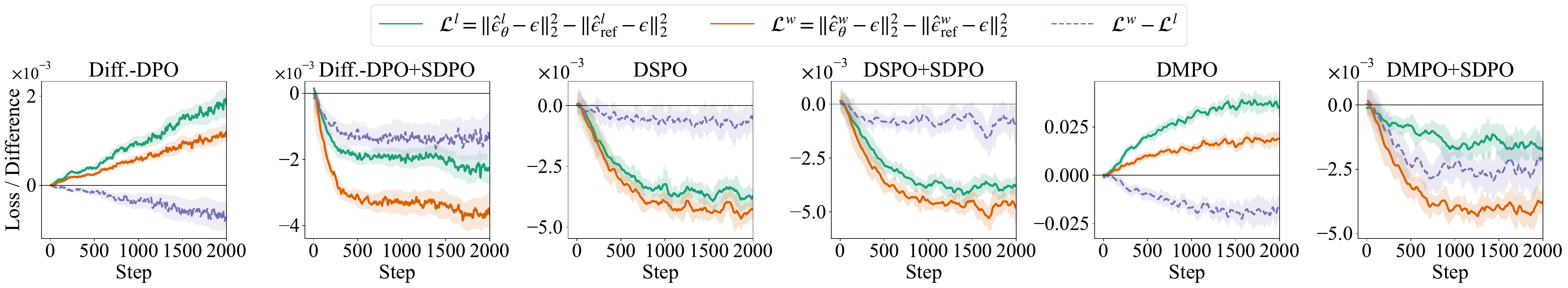}
  \caption{Training dynamics across three objectives with and without SDPO on SD 1.5.}
  \label{fig:losses_analysis}
  \vspace{-6pt}
\end{figure*}

\begin{figure*}
  \centering
  \includegraphics[width=.99\linewidth]{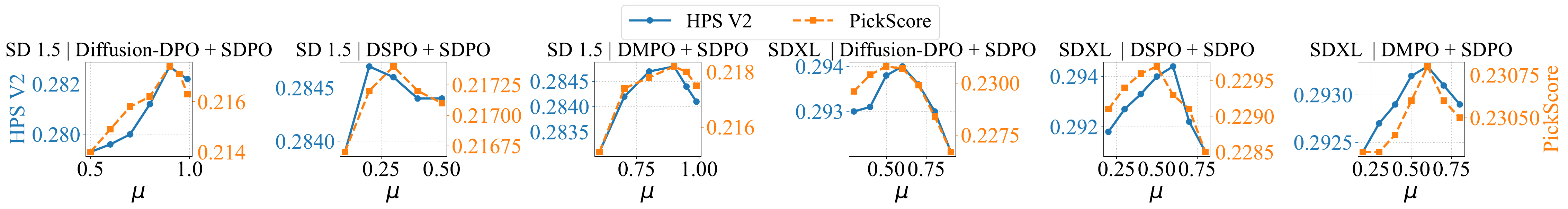}
   \caption{Sensitivity of SDPO to hyperparameter $\mu$ measured by HPS\,V2 and PickScore across SD\,1.5 and SDXL on HPS V2 prompt set.}
  \label{fig:param_analysis}
\end{figure*}

\begin{figure}
  \centering
  \includegraphics[width=.84\linewidth]{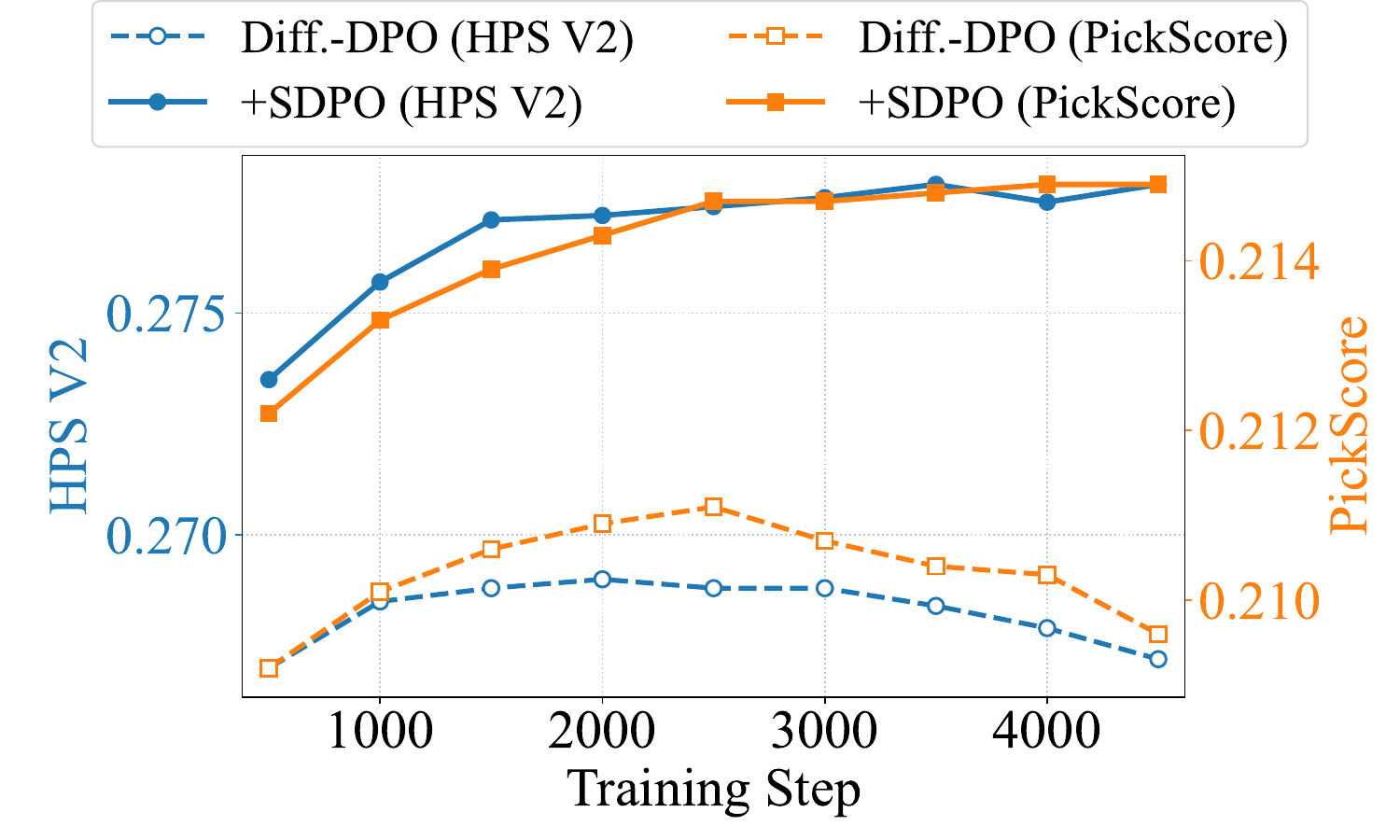}
\caption{Comparison of Diffusion–DPO with and without SDPO under longer training on SD 1.5 (Test prompts: Pick V2).}
  \label{fig:longer_training}
\end{figure}

\paragraph{Modular Ablation \& Computational Cost.}

Table~\ref{tab:unified_wpr} compares winner preserving strategies when embedded into MaPO, DPOP, and Diffusion-DPO. MaPO applies a fixed winner weight and removes the reference model, which weakens calibration of absolute error though it requires less computation and GPU memory. DPOP protects the winner through thresholded update filtering, but this rule was designed for autoregressive language models and does not fully match diffusion training dynamics. Our SDPO preserves the winner by rescaling the loser update with a safeguard coefficient $\lambda_{\text{safe}}$ selected in the output space according to directional alignment. A fixed $\lambda_{\text{safe}}$—that is, holding $\lambda_{\text{safe}}$ constant throughout training (see Table~\ref{tab:unified_wpr}, row 4)—already improves over MaPO and DPOP on PickScore and HPS, whereas allowing $\lambda_{\text{safe}}$ to adapt during training yields further gains. These improvements support the hypothesis that output–space selection of $\lambda_{\text{safe}}$ stabilizes the winner while maintaining pressure to enlarge the preference margin.

Fig.~\ref{fig:lambda_ablation} compares two mechanisms for computing the dynamic $\lambda_{\text{safe}}$: using output-space gradients and using parameter-space gradients. The output-space trajectory closely matches the parameter-space trajectory in both level and trend, indicating similar effectiveness. Because it reuses signals from the standard backward pass and only performs lightweight reductions, the output-space variant adds essentially no runtime or memory overhead relative to the Diffusion-DPO baseline. In contrast, as shown in Table~\ref{tab:unified_wpr}, computing $\lambda_{\text{safe}}$ from parameter-space gradients requires extra vector-Jacobian products and temporary buffers for per-parameter inner products, increasing training time by about 72\% and peak memory by about 11\%. 

Note that the two curves in Fig.~\ref{fig:lambda_ablation} use different values of $\mu$, which indicates that by adjusting $\mu$, the output-space scheme can approximate the behavior of parameter-space estimation. Although computing $\lambda_{\text{safe}}$ in the output space is an approximation, our empirical results show that proper tuning of $\mu$ effectively compensates for this mismatch, allowing the scaled gradients to closely match the parameter-space formulation while retaining the same training efficiency.

\paragraph{Why does SDPO generalize across DPO variants?}
Fig.~\ref{fig:losses_analysis} contrasts the training dynamics of Diff.-DPO, DSPO, and DMPO with or without SDPO. Without SDPO, $\mathcal{L}^{w}-\mathcal{L}^{l}$ decreases as expected, whereas $\mathcal{L}^{w}$ remains nondecreasing and drifts upward in Diff.-DPO and DMPO, indicating unstable optimization. With SDPO, $\mathcal{L}^{w}$ drops early and remains low, $\mathcal{L}^{l}$ declines smoothly without overshoot, and $\mathcal{L}^{w}-\mathcal{L}^{l}$ decreases steadily to a plateau. For DSPO, which already regularizes branch imbalance via its score preference objective and progressively increases the weight on the winner branch, adding SDPO causes no degradation and typically yields slightly improved reward results (cf. Table~\ref{tab:score_15}).

We observe a shared qualitative profile across the three SDPO-augmented settings: after basic rescaling, trajectories from different objectives largely overlap. $\mathcal{L}^{w}$ follows a monotone, fast-then-slow descent, $\mathcal{L}^{l}$ descends smoothly, and their gap grows in a stable manner across timesteps. This empirical regularity suggests that SDPO successfully corrects harmful update directions and magnitudes by acting on gradient geometry rather than on a particular objective form, thereby normalizing training dynamics across DPO variants, preserving the preferred branch, and stabilizing preference alignment.

\paragraph{Sensitivity of Hyperparameters.}
Fig.~\ref{fig:param_analysis} indicates that the safeguard slack $\mu$ admits broad, gently convex optima rather than sharp tuning. The resulting flat plateaus for both HPS V2 and PickScore suggest that SDPO’s behavior is governed primarily by gradient–alignment geometry in output space, rather than by the specific loss form. On SDXL, all SDPO–augmented objectives maintain near–optimal performance over a wide interval centered around $\mu\approx 0.6$, consistent with a smoother, higher–capacity landscape. 

On SD\,1.5, the more aggressive baselines (Diffusion-DPO and DMPO) benefit from a larger slack $\mu$, which further contracts the loser contribution whenever its direction conflicts with the winner. In contrast, DSPO already progressively assigns greater weight to the winner branch during training, so a smaller slack is sufficient. Practically, $\mu$ can be chosen by an early–phase heuristic: increase $\mu$ if the winner loss $\mathcal{L}^w$ drifts upward or oscillates; decrease $\mu$ in small steps if the preference margin stalls despite a stable $\mathcal{L}^w$. Following this rule of thumb, we adopt a single default on SDXL ($\mu=0.6$ for all objectives) and two defaults on SD\,1.5 ($\mu=0.9$ for Diffusion–DPO and DMPO; $\mu=0.2$ for DSPO).

\paragraph{SDPO preserves gains with longer training.}
We further examine whether the safeguarding mechanism remains effective under extended training (Fig.~\ref{fig:longer_training}). The baseline exhibits a nonmonotonic trajectory: both HPS~V2 and PickScore increase early, then plateau or decline as training proceeds. In contrast, augmenting Diffusion-DPO with SDPO yields continued gains that stabilize at a higher level, widening the gap over the baseline as steps increase. Qualitative examples in Fig.~\ref{fig:insight_combined} corroborate this trend: baseline generations develop saturation and texture artifacts under prolonged training, whereas SDPO maintains crisp structure and consistent style. Overall, SDPO sustains improvements in preference metrics without sacrificing perceptual quality during longer training. This follows from its winner-preserving constraint, which limits the influence of loser-branch gradients and ensures that the winner loss does not increase. Without SDPO, the growing contribution of the loser branch can raise the winner loss and is often accompanied by degraded visual quality.

\section{Conclusions and Limitations}
In this paper, we presented Diffusion-SDPO, a safeguarded preference optimization scheme that stabilizes DPO-style diffusion finetuning by preserving the preferred branch while improving preference matching. The method scales the loser gradient by its alignment with the winner and guarantees, to first order, that the winner’s reconstruction loss does not increase. Across SD 1.5, SDXL (both UNets), and Ovis-U1 (DiT), our method yields consistent improvements on automated preference, aesthetic, and prompt-alignment metrics with negligible computational overhead, while remaining model-agnostic, straightforward to implement, and applicable to multiple DPO variants.

However, the safeguard is derived from a first-order approximation, and its validity weakens when the loss landscape exhibits strong curvature. Moreover, the coefficient $\rho$ (cf.~Eq.~\ref{eq:grad-dot-jacobian}) used to estimate gradient alignment can be noisy or biased, which leads to imperfect scaling of loser gradients. Nevertheless, our empirical results indicate that with a suitably chosen $\mu$, this estimation is accurate enough in practice and the safeguard behaves as intended. Future work includes developing second-order or trust-region safeguards to maintain robust winner preservation under diverse training dynamics.

{
    \small
    \bibliographystyle{ieeenat_fullname}
    \bibliography{main}

@article{nc-dpo,
  title={Aligning diffusion models with noise-conditioned perception},
  author={Gambashidze, Alexander and Kulikov, Anton and Sosnin, Yuriy and Makarov, Ilya},
  journal={arXiv:2406.17636},
  year={2024}
}

@inproceedings{diffusion-dpo,
    author={Wallace, Bram and Dang, Meihua and Rafailov, Rafael and others},
    booktitle={IEEE/CVF Conference on Computer Vision and Pattern Recognition},
    title={Diffusion Model Alignment Using Direct Preference Optimization},
    year={2024},
    pages = {8228-8238},
}

@inproceedings{ddpm,
author={Ho, Jonathan and Jain, Ajay and Abbeel, Pieter},
title={Denoising diffusion probabilistic models},
year={2020},
booktitle={Advances in Neural Information Processing Systems},
pages={6840 - 6851}
}

@inproceedings{diffusion,
author = {Sohl-Dickstein, Jascha and Weiss, Eric A. and Maheswaranathan, Niru and Ganguli, Surya},
title = {Deep unsupervised learning using nonequilibrium thermodynamics},
year = {2015},
booktitle = {International Conference on Machine Learning},
pages = {2256–2265},
}

@inproceedings{dspo,
  title={{DSPO}: Direct score preference optimization for diffusion model alignment},
  author={Zhu, Huaisheng and Xiao, Teng and Honavar, Vasant G},
  booktitle={International Conference on Learning Representations},
  year={2025},
  pages={1-13}
}

@article{BT_model,
  title={Rank analysis of incomplete block designs: I. The method of paired comparisons},
  author={Bradley, Ralph Allan and Terry, Milton E},
  journal={Biometrika},
  volume={39},
  number={3/4},
  pages={324--345},
  year={1952},
  publisher={JSTOR}
}

@article{diffusion_survey,
  title={Diffusion models in vision: A survey},
  author={Croitoru, Florinel-Alin and Hondru, Vlad and Ionescu, Radu Tudor and Shah, Mubarak},
  journal={IEEE Transactions on Pattern Analysis and Machine Intelligence},
  volume={45},
  number={9},
  pages={10850-10869},
  year={2023},
  publisher={IEEE}
}

@inproceedings{sd3,
  title={Scaling rectified flow {Transformers} for high-resolution image synthesis},
  author={Esser, Patrick and Kulal, Sumith and Blattmann, Andreas and others},
  booktitle={International Conference on Machine Learning},
  year={2024},
  pages={12606-12633}
}

@misc{flux,
  title = {{FLUX.1-dev}},
  author = {Black Forest Labs},
  howpublished  = {\url{https://huggingface.co/black-forest-labs/FLUX.1-dev}},
  year = {2024}
}

@inproceedings{sdxl,
  title={{SDXL}: Improving latent diffusion models for high-resolution image synthesis},
  author={Podell, Dustin and English, Zion and Lacey, Kyle and others},
  booktitle={International Conference on Learning Representations},
  year={2024},
  pages={1-13}
}

@inproceedings{sd15,
author = {Rombach, Robin and Blattmann, Andreas and Lorenz, Dominik and Esser, Patrick and Ommer, Bjorn },
booktitle = {IEEE/CVF Conference on Computer Vision and Pattern Recognition},
title = {High-Resolution Image Synthesis with Latent Diffusion Models },
year = {2022},
pages = {10674-10685},
}

@inproceedings{cg,
author = {Dhariwal, Prafulla and Nichol, Alex},
title = {Diffusion models beat {GANs} on image synthesis},
year = {2021},
booktitle={Advances in Neural Information Processing Systems},
pages={8780-8794}
}

@inproceedings{diffusion_sde,
  title={Score-based generative modeling through stochastic differential equations},
  author={Song, Yang and Sohl-Dickstein, Jascha and Kingma, Diederik P and Kumar, Abhishek and Ermon, Stefano and Poole, Ben},
  booktitle={International Conference on Learning Representations},
  year={2022},
  pages={1-12}
}

@InProceedings{gm_sde,
  title={Score-based Generative Modeling of Graphs via the System of Stochastic Differential Equations},
  author={Jo, Jaehyeong and Lee, Seul and Hwang, Sung Ju},
  booktitle={International Conference on Machine Learning},
  pages={10362-10383},
  year={2022},
}

@inproceedings{edm,
    author = {Karras, Tero and Aittala, Miika and Laine, Samuli and Aila, Timo},
    title = {Elucidating the design space of diffusion-based generative models},
    year = {2022},
    booktitle = {Advances in Neural Information Processing Systems},
    pages = {26565 - 26577}
}

@inproceedings{rectified_flow,
  title={Flow straight and fast: Learning to generate and transfer data with rectified flow},
  author={Liu, Xingchao and Gong, Chengyue and Liu, Qiang},
  booktitle={International Conference on Learning Representations},
  year={2023},
  pages={1-15}
}

@inproceedings{flow_matching,
  title={Flow matching for generative modeling},
  author={Lipman, Yaron and Chen, Ricky TQ and Ben-Hamu, Heli and Nickel, Maximilian and Le, Matt},
  booktitle={International Conference on Learning Representations},
  year={2023},
  pages={1-13}
}

@article{cfg,
  title={Classifier-free diffusion guidance},
  author={Ho, Jonathan and Salimans, Tim},
  journal={arXiv:2207.12598},
  year={2022}
}

@misc{google2025_nano_banana,
  title        = {Introducing Gemini 2.5 Flash Image (aka nano-banana)},
  author       = {{Google}},
  year         = {2025},
  howpublished          = {\url{https://developers.googleblog.com/en/introducing-gemini-2-5-flash-image/}}
}

@inproceedings{llm_dpo,
    author = {Rafailov, Rafael and Sharma, Archit and Mitchell, Eric and Ermon, Stefano and Manning, Christopher D. and Finn, Chelsea},
    title = {Direct preference optimization: your language model is secretly a reward model},
    year = {2023},
    booktitle = {Advances in Neural Information Processing Systems},
    pages={53728-53741}
}

@article{balanceddpo,
  title={{BalancedDPO}: Adaptive multi-metric alignment},
  author={Tamboli, Dipesh and Chakraborty, Souradip and Malusare, Aditya and Banerjee, Biplab and Bedi, Amrit Singh and Aggarwal, Vaneet},
  journal={arXiv:2503.12575},
  year={2025}
}

@article{lpo,
  title={Linear Preference Optimization: Decoupled Gradient Control via Absolute Regularization},
  author={Wang, Rui and Sun, Qianguo and Song, Chao and Wu, Junlong and Chen, Tianrong and Zeng, Zhiyun and Li, Yu},
  journal={arXiv:2508.14947},
  year={2025}
}

@inproceedings{mapo,
  title={Margin-aware preference optimization for aligning diffusion models without reference},
  author={Hong, Jiwoo and Paul, Sayak and Lee, Noah and Rasul, Kashif and Thorne, James and Jeong, Jongheon},
  booktitle={International Conference on Learning Representations Workshop},
  year={2025}
}

@article{orthogonal-dpo,
  title={Orthogonal Finetuning for Direct Preference Optimization},
  author={Yang, Chenxu and Jia, Ruipeng and Gu, Naibin and others},
  journal={arXiv:2409.14836},
  year={2024}
}

@article{bounded-dpo,
  title={{Rethinking DPO}: The Role of Rejected Responses in Preference Misalignment},
  author={Cho, Jay Hyeon and Oh, JunHyeok and Kim, Myunsoo and Lee, Byung-Jun},
  journal={arXiv:2506.12725},
  year={2025}
}

@article{llm_survey,
  title={A survey of large language models},
  author={Zhao, Wayne Xin and Zhou, Kun and Li, Junyi and others},
  journal={arXiv:2303.18223},
  year={2023}
}

@inproceedings{calibrated-dpo,
  title={{Cal-DPO}: Calibrated direct preference optimization for language model alignment},
  author={Xiao, Teng and Yuan, Yige and Zhu, Huaisheng and Li, Mingxiao and Honavar, Vasant G},
  booktitle={Advances in Neural Information Processing Systems},
  pages={114289-114320},
  year={2024}
}

@inproceedings{robust-dpo,
    author = {Chowdhury, Sayak Ray and Kini, Anush and Natarajan, Nagarajan},
    title = {Provably robust {DPO}: aligning language models with noisy feedback},
    year = {2024},
    pages={42258 - 42274},
    booktitle = {International Conference on Machine Learning},
}

@inproceedings{chi2-po,
  title={Correcting the mythos of KL-regularization: Direct alignment without overparameterization via Chi-squared preference optimization},
  author={Huang, Audrey and Zhan, Wenhao and Xie, Tengyang and Lee, Jason D and Sun, Wen and Krishnamurthy, Akshay and Foster, Dylan J},
  booktitle={International Conference on Learning Representations},
  year={2025},
  pages={1-16}
}

@article{dpop,
  title={{Smaug}: Fixing failure modes of preference optimisation with dpo-positive},
  author={Pal, Arka and Karkhanis, Deep and Dooley, Samuel and Roberts, Manley and Naidu, Siddartha and White, Colin},
  journal={arXiv:2402.13228},
  year={2024}
}

@article{mappo,
  title={{MaPPO}: Maximum a Posteriori Preference Optimization with Prior Knowledge},
  author={Lan, Guangchen and Zhang, Sipeng and Wang, Tianle and others},
  journal={arXiv:2507.21183},
  year={2025}
}

@article{see-dpo,
    title={{SEE}-{DPO}: Self Entropy Enhanced Direct Preference Optimization},
    author={Shivanshu Shekhar and Shreyas Singh and Tong Zhang},
    journal={Transactions on Machine Learning Research},
    issn={2835-8856},
    year={2025},
}

@article{dmpo,
  title={Divergence Minimization Preference Optimization for Diffusion Model Alignment},
  author={Li, Binxu and Xu, Minkai and Dang, Meihua and Ermon, Stefano},
  journal={arXiv:2507.07510},
  year={2025}
}

@inproceedings{rlhf,
author = {Christiano, Paul F. and Leike, Jan and Brown, Tom B. and Martic, Miljan and Legg, Shane and Amodei, Dario},
title = {Deep reinforcement learning from human preferences},
year = {2017},
booktitle={Advances in Neural Information Processing Systems},
pages={4302-4310}
}

@inproceedings{diffusion-kto,
    author = {Li, Shufan and Kallidromitis, Konstantinos and Gokul, Akash and Kato, Yusuke and Kozuka, Kazuki},
    title = {Aligning diffusion models by optimizing human utility},
    year = {2025},
    booktitle={Advances in Neural Information Processing Systems},
    pages={24897 - 24925}
}

@article{ovis-u1,
  title={{Ovis-U1} Technical Report},
  author={Wang, Guo-Hua and Zhao, Shanshan and Zhang, Xinjie and others},
  journal={arXiv:2506.23044},
  year={2025}
}

@inproceedings{pap,
    author = {Kirstain, Yuval and Polyak, Adam and Singer, Uriel and Matiana, Shahbuland and Penna, Joe and Levy, Omer},
    title = {{Pick-a-Pic}: An open dataset of user preferences for text-to-image generation},
    year = {2023},
    booktitle={Advances in Neural Information Processing Systems},
    pages = {36652 - 36663}
}

@article{hpsv2,
  title={{Human Preference Score v2}: A solid benchmark for evaluating human preferences of text-to-image synthesis},
  author={Wu, Xiaoshi and Hao, Yiming and Sun, Keqiang and Chen, Yixiong and Zhu, Feng and Zhao, Rui and Li, Hongsheng},
  journal={arXiv:2306.09341},
  year={2023}
}

@article{partiprompts,
  title={Scaling autoregressive models for content-rich text-to-image generation},
  author={Yu, Jiahui and Xu, Yuanzhong and Koh, Jing Yu and others},
  journal={arXiv:2206.10789},
  year={2022}
}

@inproceedings{clip,
  title={Learning transferable visual models from natural language supervision},
  author={Radford, Alec and Kim, Jong Wook and Hallacy, Chris and others},
  booktitle={International Conference on Machine Learning},
  pages={8748-8763},
  year={2021}
}

@misc{laion-aesthetics,
  author       = {Schuhmann, Christoph and Vencu, Romain and Beaumont, Romai and others},
  title        = {{LAION-Aesthetics}: Predicting the Aesthetic Quality of Images},
  howpublished = {\url{https://laion.ai/blog/laion-aesthetics/}},
  year         = {2022}
}

@inproceedings{imagereward,
author={Xu, Jiazheng and Liu, Xiao and Wu, Yuchen and others},
title={{ImageReward}: Learning and evaluating human preferences for text-to-image generation},
year={2024},
booktitle={Advances in Neural Information Processing Systems},
pages={15903-15935}
}

@inproceedings{unet,
  title={{U-net}: Convolutional networks for biomedical image segmentation},
  author={Ronneberger, Olaf and Fischer, Philipp and Brox, Thomas},
  booktitle={International Conference on Medical Image Computing and Computer-Assisted Intervention},
  pages={234-241},
  year={2015}
}

@inproceedings{geneval,
author = {Ghosh, Dhruba and Hajishirzi, Hannaneh and Schmidt, Ludwig},
title = {{GENEVAL}: An object-focused framework for evaluating text-to-image alignment},
booktitle={Advances in Neural Information Processing Systems},
year = {2023},
page = {52132-52152},
}

@article{dpg_bench,
  title={{EllA}: Equip diffusion models with {LLM} for enhanced semantic alignment},
  author={Hu, Xiwei and Wang, Rui and Fang, Yixiao and Fu, Bin and Cheng, Pei and Yu, Gang},
  journal={arXiv:2403.05135},
  year={2024}
}

@article{imgedit,
  title={{Imgedit}: A unified image editing dataset and benchmark},
  author={Ye, Yang and He, Xianyi and Li, Zongjian and others},
  journal={arXiv:2505.20275},
  year={2025}
}

@article{gedit,
  title={{Step1X-Edit}: A practical framework for general image editing},
  author={Liu, Shiyu and Han, Yucheng and Xing, Peng and others},
  journal={arXiv:2504.17761},
  year={2025}
}

@inproceedings{dit,
author = {Peebles, William and Xie, Saining },
booktitle = {IEEE/CVF International Conference on Computer Vision},
title = {Scalable Diffusion Models with {Transformers}},
year = {2023},
pages = {4172-4182},
}

@article{qwen_vl_max,
  title={{Qwen-VL}: A Versatile Vision-Language Model for Understanding, Localization, Text Reading, and Beyond},
  author={Jinze Bai and Shuai Bai and Shusheng Yang and others},
  journal={arXiv:2308.12966},
  year={2023}
}
}

\appendix

\section{Second-Order Considerations of SDPO} 
\label{sec:second_order_appendix}

Our theoretical guarantee for Diffusion-SDPO is explicitly first order: it controls the sign of the linear term in the Taylor expansion of the winner loss. In practice, the true change in $\mathcal L^w$ after an update is
\begin{equation}
\Delta \mathcal L^w 
= \nabla_\theta {\mathcal L^w}^\top \Delta \theta 
+ \tfrac{1}{2}\Delta \theta^\top H^w \Delta \theta 
+ \mathcal{O}(\|\Delta \theta\|^3),
\end{equation}
where $H^w$ is the Hessian of $\mathcal L^w$ and $\Delta \theta$ is the parameter update induced by the SDPO objective. The analysis in main text ensures that the \emph{first-order} term $\nabla_\theta {\mathcal L^w}^\top \Delta \theta$ is nonpositive under the safe choice of $\lambda$. However, the quadratic term $\tfrac{1}{2}\Delta \theta^\top H^w \Delta \theta$ can in principle be positive when local curvature is large, so $\mathcal L^w$ might still increase slightly when the step is not infinitesimal. In other words, SDPO controls the \emph{direction} of the update with respect to the gradient of $\mathcal L^w$, while the Hessian governs how quickly the loss can bend back upward along that direction.

The role of the slack parameter $\mu$ to compute $\lambda_\text{safe}$ can then be understood as an implicit trust-region style safeguard. Recall that the parameter-space analysis yields an upper bound on $\lambda$, and our output-space scheme uses
\begin{equation}
\lambda_{\text{safe}} = \frac{(1-\mu)\|g^w\|_2^2}{g^{w\top} g^l},
\end{equation}
which shrinks the allowable contribution of the loser branch whenever $g^{w\top} g^l > 0$. Multiplying by $(1-\mu)$ reduces the contribution of the loser gradient in the update. In the regime where $\nabla_\theta \mathcal L^w{}^\top \nabla_\theta \mathcal L^l > 0$, the first-order change in Eq.~\ref{eq:delta-Lw-appendix},
\begin{equation}
\Delta \mathcal L^w \!\!\approx\!\! \nabla_\theta {\mathcal L^w}^\top\!\!\Delta \theta \!=\! \!-\eta\Big(\!\|\nabla_\theta \mathcal L^w\|_2^2 \!-\! \lambda\nabla_\theta {\mathcal L^w}^\top\!\nabla_\theta \mathcal L^l\!\Big),
\label{eq:delta-Lw-appendix}
\end{equation}
is monotonically increasing as a function of $\lambda$ (note that both $\lambda$ and $\eta$ are $>0$), so replacing $\lambda$ by $(1-\mu)\lambda$ makes $\Delta \mathcal L^w$ strictly smaller. 

At the same time, we can make the dependence on the loser direction explicit. The SDPO update can be written as
\begin{equation}
\Delta \theta(\lambda)
\!=\! -\eta\big(\nabla_\theta \mathcal L^w \!\!\!-\!\! \lambda \nabla_\theta \mathcal L^l\big)
\!=\! \underbrace{-\eta\,\nabla_\theta \mathcal L^w}_{\Delta\theta(0)}
\!\!+\!\!\!\!\! \underbrace{\eta\lambda\,\nabla_\theta \mathcal L^l}_{\text{loser-induced part}}\!\!\!\!.
\label{eq:sdpo-decomp}
\end{equation}
The term $\eta\lambda\,\nabla_\theta \mathcal L^l$ is the $\lambda$-dependent increment of the update along the loser direction. Replacing $\lambda$ by $(1-\mu)\lambda$ scales this increment to $\eta(1-\mu)\lambda\,\nabla_\theta \mathcal L^l$, so its norm is multiplied by $(1-\mu)$ while the baseline part $\Delta\theta(0)$ is unchanged. In particular, the portion of the step norm that is directly attributable to the loser branch contracts linearly with $(1-\mu)$. The quadratic term then decomposes as
\begin{equation}
\begin{aligned}
&\Delta \theta(\lambda)^\top H^w \Delta \theta(\lambda)
= \Delta\theta(0)^\top H^w \Delta\theta(0) \\
&\!\!\!\!\!\!\quad + 2\eta\lambda\, (\nabla_\theta \mathcal L^l)^\top H^w \Delta\theta(0)
\!+\! \eta^2\lambda^2\, (\nabla_\theta \mathcal L^l)^\top \!H^w \nabla_\theta \mathcal L^l\!,
\end{aligned}
\label{eq:second-order-decomp}
\end{equation}
where the last two terms capture the curvature contribution associated with the loser direction. Under a mild local spectral bound $\|H^w\|_2 \leq \Lambda$, we can control the quadratic term via
\begin{equation}
\bigl|v^\top H^w v\bigr|
\leq \|H^w\|_2 \,\|v\|_2^2
\leq \Lambda \,\|v\|_2^2,
\end{equation}
for any vector $v$. Applying this with $v = \Delta\theta\bigl((1-\mu)\lambda\bigr)$ and using the decomposition in Eq.~\eqref{eq:sdpo-decomp} yields
\begin{align}
    &\Bigl|\tfrac{1}{2}\Delta \theta\bigl((1-\mu)\lambda\bigr)^\top H^w \Delta \theta\bigl((1-\mu)\lambda\bigr)\Bigr| \\
    &\leq \tfrac{1}{2}\Lambda \bigl\|\Delta \theta\bigl((1-\mu)\lambda\bigr)\bigr\|_2^2 \\
    &= \tfrac{1}{2}\Lambda \bigl\|\Delta\theta(0) + \eta(1-\mu)\lambda\,\nabla_\theta \mathcal L^l\bigr\|_2^2 \\
    &\leq \tfrac{1}{2}\Lambda \Bigl(\bigl\|\Delta\theta(0)\bigr\|_2 
        + \eta(1-\mu)|\lambda|\,\bigl\|\nabla_\theta \mathcal L^l\bigr\|_2\Bigr)^2 \\
    &\leq \tfrac{1}{2}\Lambda \Bigl(\bigl\|\Delta\theta(0)\bigr\|_2 
        + \eta|\lambda|\,\bigl\|\nabla_\theta \mathcal L^l\bigr\|_2\Bigr)^2.
    \label{eq:contracted-bound-final}
\end{align}
The last line coincides with the corresponding spectral-norm bound obtained for $\Delta\theta(\lambda)$ before contraction. Since $(1-\mu)|\lambda| \leq |\lambda|$ for $\mu\in[0,1]$, the above inequalities show that \textit{the curvature contribution under the contracted update is controlled by an upper bound that is no larger, and typically strictly smaller, than the original one}. We therefore claim that SDPO shrinks the worst-case curvature impact in the loser direction, making it less likely that higher-order effects overturn the negative first-order change.

This perspective also clarifies the interaction between SDPO and the base optimizer. In our experiments we use \textit{small learning rates} and \textit{large gradient accumulation}, so the effective $\|\Delta \theta\|$ per update is modest even for high-capacity backbones. In this regime, the second-order contribution behaves as a small perturbation around the controlled first-order term. When combined with the slack contraction, this explains why we observe almost monotone or gently decreasing $\mathcal L^w$ trajectories, rather than the oscillatory behavior that would indicate strong curvature dominating the dynamics.

Empirically, we can also see second-order effects indirectly through the sensitivity of SDPO to $\mu$. If curvature induced large and frequent violations of the first-order approximation, performance would vary sharply with small changes in $\mu$ because the balance between the linear and quadratic terms would be fragile. Instead, we observe broad plateaus where both HPS V2 and PickScore remain near-optimal over wide intervals of $\mu$. This suggests that, in the regions visited during training, the winner loss is reasonably well approximated by its first-order geometry along SDPO updates, and the quadratic term acts as a small correction rather than a dominant force.

Finally, the approximate match between parameter-space and output-space trajectories offers further evidence that higher-order interactions are not pathological in practice. The parameter-space variant implicitly incorporates more of the true curvature structure, yet its behavior can be closely reproduced by the cheaper output-space scheme with a suitably chosen slack. This indicates that most of the relevant geometry is already captured by the alignment between $g^w$ and $g^l$, and that the remaining second-order discrepancy can be effectively absorbed into a scalar safety margin. A fully second-order SDPO variant that explicitly constrains $\Delta \theta^\top H^w \Delta \theta$ would be an interesting direction for future work, but our results suggest that the present first-order safeguard, combined with the slack $\mu$, already offers a favorable trade-off between theoretical control, computational cost, and empirical stability.

\begin{table*}[t]
\centering
\setlength{\tabcolsep}{2pt}
\caption{Full reward and structured alignment results for FLUX.1-dev. We report automatic rewards (PickScore, HPS~V2, LAION Aesthetics, CLIP, and ImageReward) as well as structured alignment metrics (GenEval and DPG-Bench) for the base model and its preference-aligned variants. The row marked with $\S$ corresponds to Diff.-DPO trained with a $0.01\times$ learning rate.}
\begin{tabular}{l|ccccc|cc}
    \hline
    \textbf{Method} 
    & \textbf{PickScore($\uparrow$)} 
    & \textbf{HPS~V2($\uparrow$)} 
    & \textbf{Aesthetics($\uparrow$)} 
    & \textbf{CLIP($\uparrow$)} 
    & \textbf{ImageReward($\uparrow$)} 
    & \textbf{GenEval($\uparrow$)} 
    & \textbf{DPG-Bench($\uparrow$)} \\
    \hline
    FLUX.1-dev   & 0.2252 & 0.2889 & 6.0735 & 0.3498 & 0.9916 & 0.66 & 85.75 \\
    Diff.-DPO            & 0.1998 & 0.2467 & 5.2857 & 0.2573 & -0.7788 & 0.23  & 58.97 \\
    Diff.-DPO$^{\S}$   & 0.2253 & 0.2881 & 6.0701 & 0.3501 & 0.9938 & 0.66 & 85.78 \\
    Diff.-DPO + SDPO     & \bf 0.2314 & \bf 0.2991 & \bf 6.1910 & \bf 0.3566 & \bf 1.1108 & \bf 0.71 & \bf 86.36 \\
    \hline
\end{tabular}
\label{tab:flux_papv2_full}
\end{table*}

\section{Extension: SDPO on FLUX.1-dev}

\paragraph{Model.}
To further test the generality of SDPO, we also apply it to FLUX.1-dev~\cite{flux}, a 12B-parameter rectified-flow DiT~\cite{dit} model designed for text-to-image generation. FLUX.1-dev operates in latent space with a flow-matching objective rather than the noise-prediction objective used in SD~1.5 and SDXL, but it still exposes a time-conditioned image generator whose outputs can be scored by the same preference metrics. In our experiments, we start from the publicly released FLUX.1-dev checkpoint and keep the official tokenizer, text encoder, VAE, and sampling configuration unchanged. We integrate Diffusion-DPO and SDPO at the loss level only, treating the model’s velocity (or denoising) prediction at each time step as the output on which we define the per-step winner and loser residuals.

\paragraph{Datasets.}
For DPO training on FLUX.1-dev, we construct an in-house preference dataset of 186K high-quality DPO examples. The images and texts cover a broad range of domains (portraits, everyday objects, indoor and outdoor scenes, concept art, products, etc.), making the corpus suitable for general-purpose alignment rather than a narrow style or category. To improve the textual side of supervision, we apply a vision–language model (Qwen-VL-Max~\cite{qwen_vl_max}) to re-caption short, noisy, or underspecified prompts, yielding richer and more semantically faithful descriptions while preserving the original user intent.

\paragraph{Training Details.}
We finetune FLUX.1-dev with Diffusion-DPO and with our SDPO-augmented variant using TorchTitan in a hybrid sharded data-parallel (HSDP) configuration. Within each 8-GPU group we apply fully sharded data parallelism (FSDP) to partition model parameters, and we use 2-way data parallelism across groups, resulting in 16 NVIDIA A100 GPUs in total. Training is performed at a resolution of $1024\times 1024$ with a global batch size of 512 and a learning rate of $1\times 10^{-5}$. For the DPO temperature, we set $\beta_{\text{DPO}} = 1000$. For SDPO, we choose a slack of $\mu = 0.99$, which empirically yields effective safe scaling coefficients $\lambda_{\text{safe}} \approx 0.1$ at typical training steps. Under this configuration, finetuning completes in approximately three days on 16 A100 GPUs.

\paragraph{Evaluation.}
We evaluate FLUX.1-dev using the Pick-a-Pic~V2~\cite{pap} prompt sets with automatic reward metrics (PickScore~\cite{pap}, HPS~V2~\cite{hpsv2}, LAION Aesthetics~\cite{laion-aesthetics}, CLIP~\cite{clip}, and ImageReward~\cite{imagereward}), and additionally report its structured alignment performance on GenEval~\cite{geneval} and DPG-Bench~\cite{dpg_bench}. During inference, we adopt a 50-step sampler and follow the FLUX.1-dev model card by using a classifier-free guidance scale of $3.5$ for all variants to ensure a fair comparison.

\paragraph{Results.}
Table~\ref{tab:flux_papv2_full} reveals a clear three-regime behavior on FLUX.1-dev. At the default learning rate ($1\times 10^{-5}$), naive Diffusion-DPO catastrophically degrades the model: all five reward metrics drop, and both GenEval and DPG-Bench scores collapse. This suggests that on a high-capacity rectified-flow backbone, aggressively enlarging the preference margin without any safeguard on the winner branch can drive the parameters far outside the local basin of the pretrained solution. 

When we instead reduce the learning rate by a factor of $0.01$ (Diff.-DPO$^{\S}$ with learning rate $1\times 10^{-7}$), the collapse disappears and all metrics revert to nearly the base FLUX.1-dev level, indicating that the optimization has become so conservative that it behaves almost like an identity map and fails to extract meaningful gains from the preference supervision. 

Taken together, these two extremes highlight a practical tension when directly applying Diffusion-DPO to large rectified-flow models: in naive settings, one tends to end up either in an aggressive regime that destabilizes the pretrained generator, or in an overly conservative regime where the model remains effectively unchanged.

In contrast, Diff.-DPO + SDPO improves over the base model on \emph{all} reported metrics while keeping the original, high learning rate. PickScore, HPS~V2, Aesthetics, CLIP, and ImageReward all increase, and both GenEval and DPG-Bench scores also rise, indicating that SDPO not only enhances perceptual quality but also strengthens structured text–image alignment.

\section{Other Experimental Results}
\label{sec:full_results_appendix}

We report the full reward score comparisons on Pick-a-Pic~V2, HPS~V2, and PartiPrompts in Tables~\ref{tab:score_15_full} and~\ref{tab:score_xl_full}, and summarize the win-rate comparison of SDXL models in Table~\ref{tab:hpsv2_winrate_sdxl}. Across all datasets, adding SDPO on top of Diffusion-DPO, DSPO, or DMPO consistently shifts the metric profile in the same direction as in the main text: PickScore, HPS~V2, and ImageReward improve while aesthetic scores are preserved or slightly enhanced, indicating that SDPO strengthens preference alignment without paying a quality penalty. 

\begin{figure*}
  \centering
  \includegraphics[width=.9\linewidth]{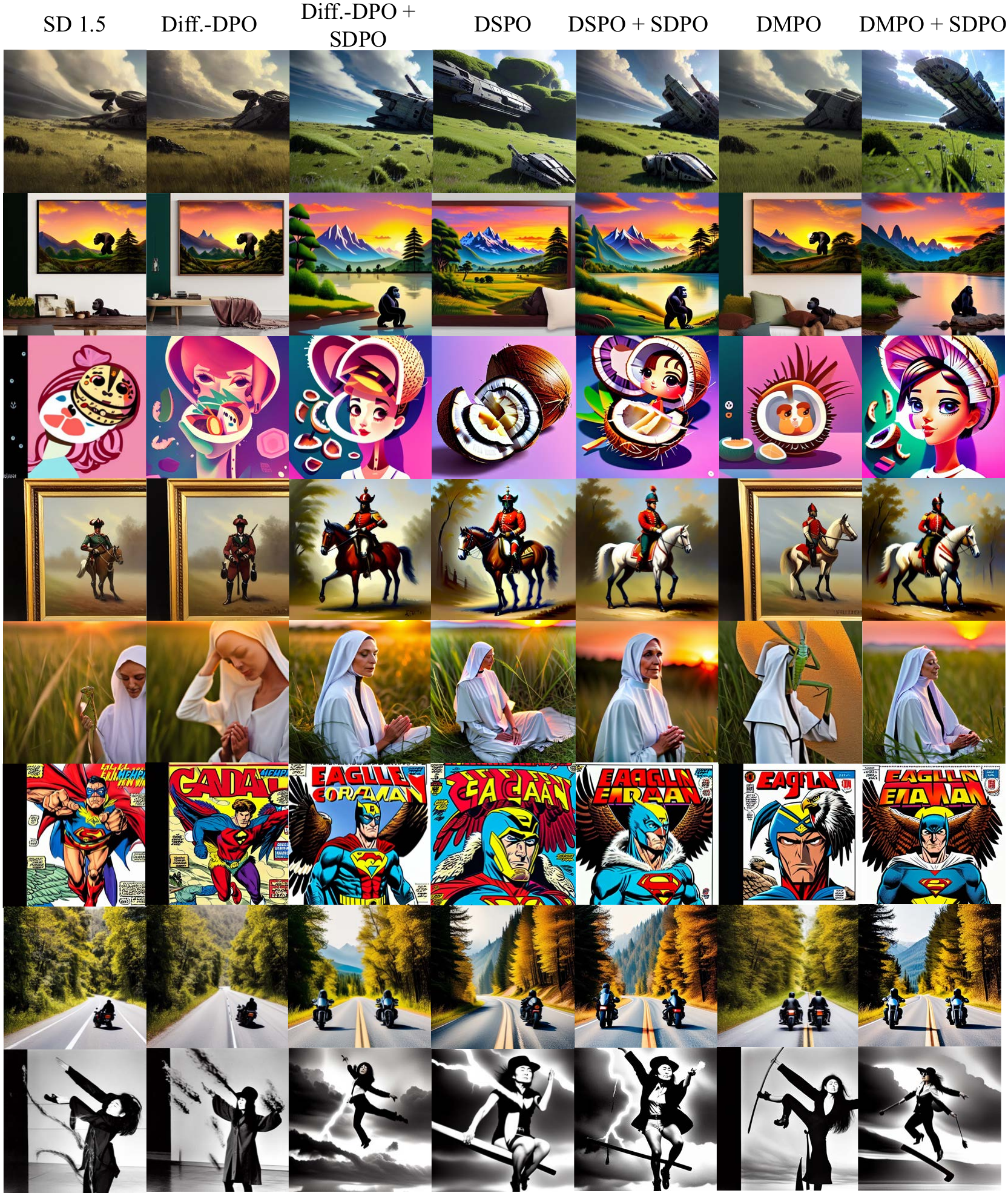}
\caption{Qualitative comparison of images generated by different methods using SD 1.5. Prompt: \textit{1) A hyper-realistic landscape from a Neil Blomkamp film featuring a crashed spaceship, detailed grass, and a photorealistic sky. 2) A landscape featuring mountains, a valley, sunset light, wildlife and a gorilla, reminiscent of Bob Ross's artwork. 3) A stylized portrait featuring sliced coconut, electronics, and AI in a cartoonish cute setting with a dramatic atmosphere. 4) A tonalist painting of a bipedal pony creature soldier. 5) A praying mantis nun in a grassy field during sunset. 6) A comic book cover featuring a superhero named "Eagle Man" with an eagle mask and wing logo, resembling a traditional comic book cover. 7) Two motorcycles sit on the side of a secluded road. 8) Yoko Ono flying on a broomstick with lightning in the skies.}}
\label{fig:images_sd15_appendix}
\end{figure*}

\begin{figure*}
  \centering
  \includegraphics[width=.95\linewidth]{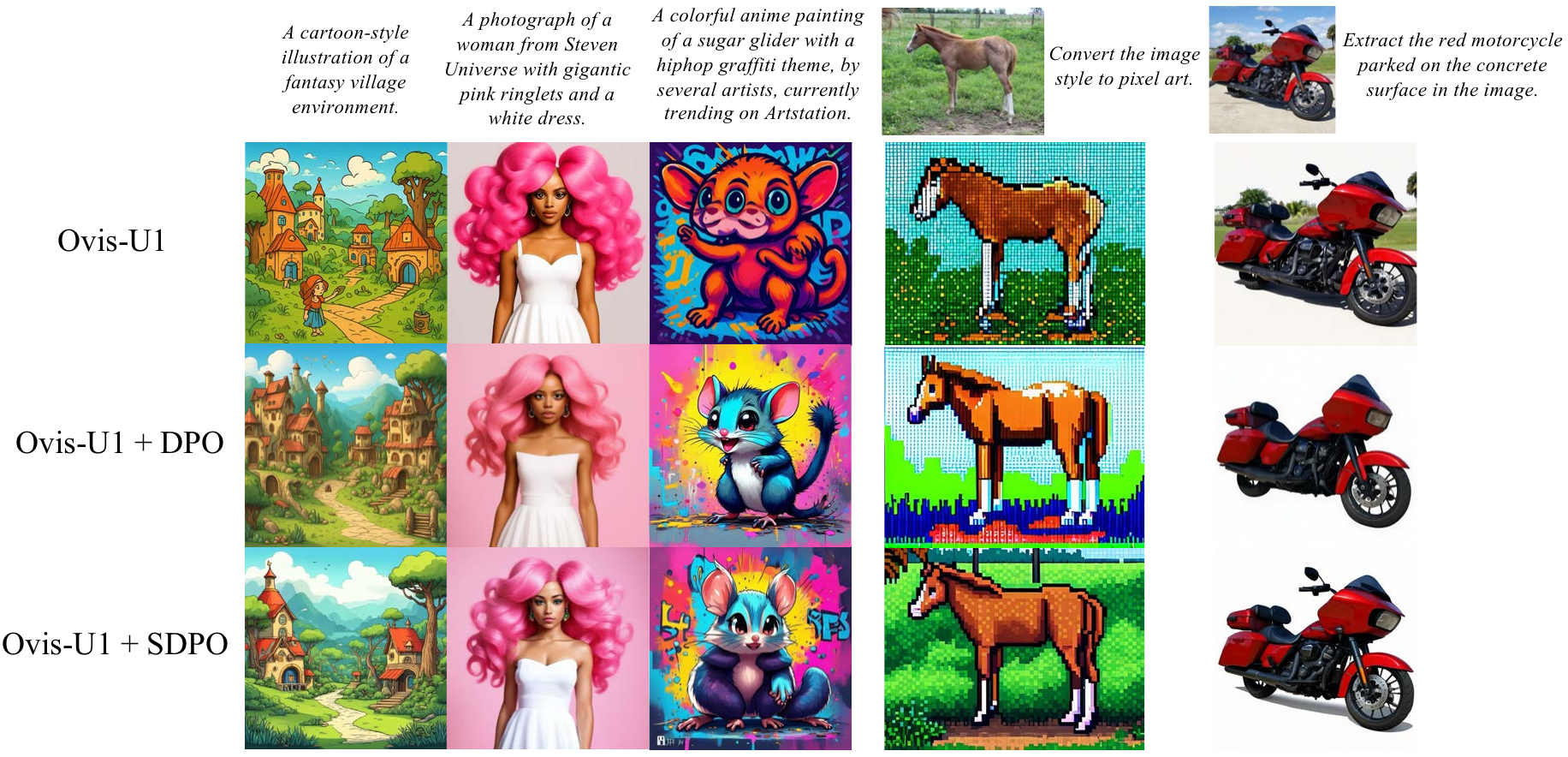}
\caption{Qualitative comparison of images generated by Ovis-U1~\cite{ovis-u1} and its finetuned variants. Results are reported for three variants: the base, finetuned with DPO, and finetuned with our SDPO.}
\label{fig:images_ovis_u1}
\end{figure*}

\begin{figure*}
  \centering
  \includegraphics[width=.95\linewidth]{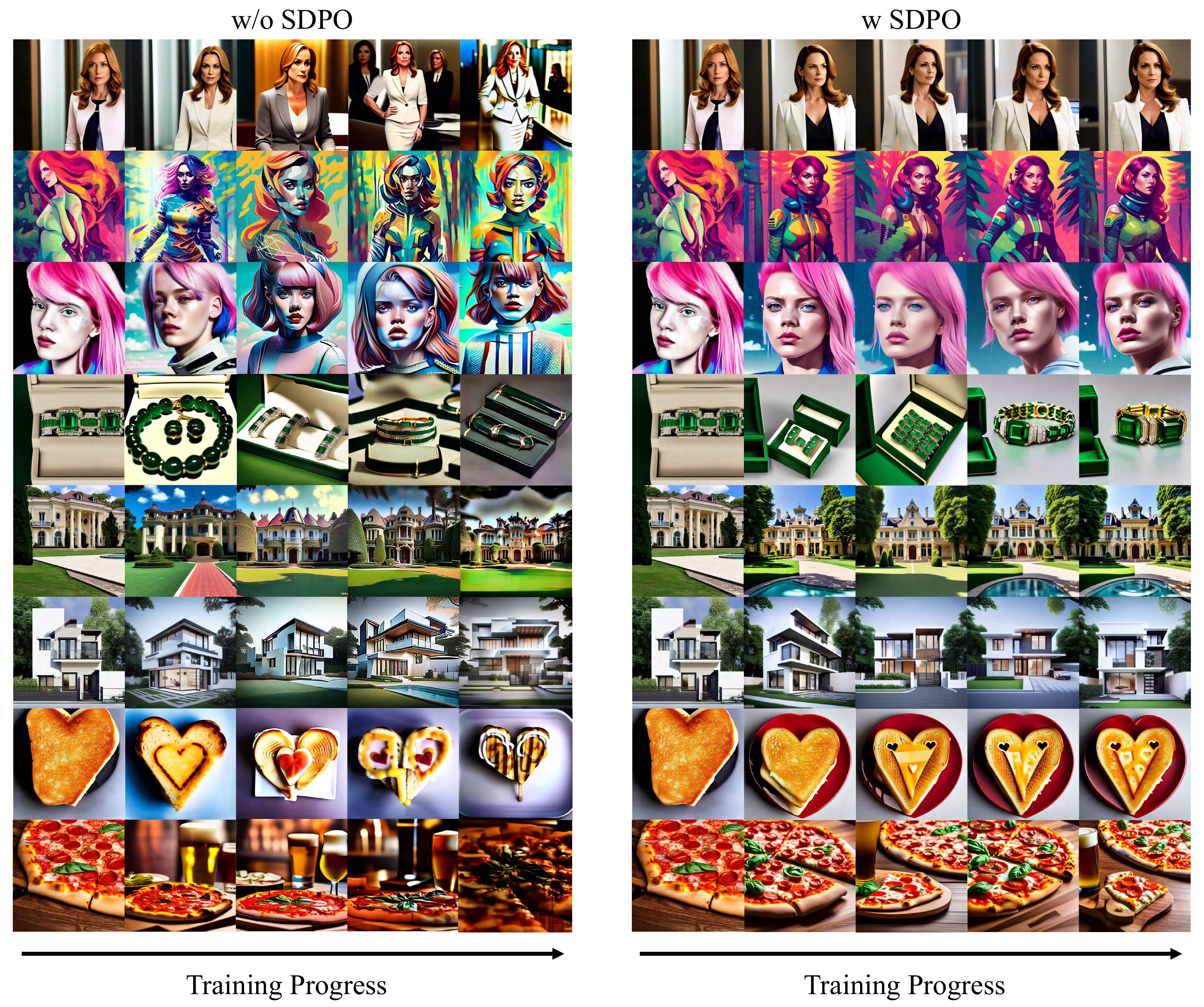}
    \caption{Qualitative comparison of generations from preference-aligned models \emph{without} SDPO (left) and \emph{with} SDPO (right). Each row corresponds to a fixed prompt, and within each panel the training step increases from left to right. Without SDPO, the baseline gradually drifts and produces saturated, distorted, or low-fidelity images as the loser gradients degrades the winner branch. In contrast, SDPO keeps the winner branch stable, preserving structure, style, and prompt alignment even at later training steps. \textit{Prompts: 1) a career woman on suits. 2) portrait of a beautiful female space warrior in a dense forest, by fiona staples, bold colors, dynamic composition, bright saturated hues, strong constrasts, vibrant, energetic, elaborate, hyper detailed, visually stunning and captivating art style. 3) pink-haired woman looking straight ahead, full lips, white military clothing with small red details, blue sky with blurred clouds, Chromatic Aberration, Geometric Shape, Photorealistic, Cosmic, Detailed, Bloom, masterpiece, best quality, extremely detailed CG unity 8k wallpaper, landscape, 3D Digital Paintings, award winning photography, Photorealistic, trending on artstation, trending on CGsociety, Intricate, High Detail, dramatic, high quality lighting, vivid anime color. 4) A set of emerald bracelets in green in a display box at the auction, uplight, very realist, very detailed, highest resolution, hyper realistic. 5) big mansion in the daytime. 6) two story house, contemporary minimalistic architecture, photorealistic rendering. 7) grilled cheese in the shape of a heart. 8) close up of Italian pizza margherita, a glass of fresh beer, candle light, polished wood table, UHD, high quality, high detail, ultra definition, high octane render, Style anime \#cibo \#pizza.}}
    \label{fig:step_images}
\end{figure*}

Additional qualitative examples in Fig.~\ref{fig:images_sd15_appendix} and Fig.~\ref{fig:images_ovis_u1} further illustrate this effect on both UNet and DiT backbones, showing sharper details, more faithful layouts, and fewer artifacts compared to their non-SDPO counterparts. Finally, Fig.~\ref{fig:step_images} visualizes the temporal dynamics: as training progresses, baseline Diffusion-DPO drifts toward over-saturated or distorted images, illustrating how the winner loss can keep increasing even while the winner-loser margin becomes more negative when optimization focuses purely on enlarging the margin. In contrast, SDPO maintains coherent structure and style across steps, thanks to its winner-preserving update rule and the stabilizing effect of the safe scaling $\lambda_{\text{safe}}$. Taken together, these quantitative and qualitative results support our central claim that SDPO acts as a plug-in, geometry-aware safeguard that consistently improves preference metrics while maintaining, and in many cases enhancing, the visual quality of diffusion generations.

\begin{table*}[t]
\centering
\footnotesize
\setlength{\tabcolsep}{3.5pt}
\caption{Full results of reward score comparison on Pick-a-Pic V2, HPS V2, and PartiPrompts using SD 1.5. $^\dagger$: results from our implementation due to the lack of official code.}
\begin{tabular}{l|l|ccccc}
    \hline
    \textbf{Dataset} & \textbf{Method} 
    & \textbf{PickScore($\uparrow$)} 
    & \textbf{\hphantom{0} HPS($\uparrow$) \hphantom{0}} 
    & \textbf{Aesthetics($\uparrow$)} 
    & \textbf{\hphantom{0} CLIP($\uparrow$) \hphantom{0}} 
    & \textbf{Image Reward($\uparrow$)} \\
    \hline
    \multirow{11}{*}{Pick V2} 
    & SD 1.5 & 0.2073 & 0.2651 & 5.3907 & 0.3299 & -0.1376 \\
    & SFT & 0.2128 & 0.2765 & 5.6888 & 0.3408 & 0.5767 \\
    & Diff.-KTO & 0.2126 & 0.2766 & 5.6288 & 0.3420 & 0.5593 \\
    & MaPO$^\dagger$ & 0.2097 & 0.2702 & 5.5572 & 0.3365 & 0.2435 \\
    & DPOP$^\dagger$ & 0.2119 & 0.2726 & 5.5688 & 0.3389 & 0.3259 \\ 
    \cdashline{2-7}
    & Diff.-DPO & 0.2109 & 0.2690 & 5.4958 & 0.3357 & 0.1020 \\
    & \hphantom{0} + \textbf{SDPO} & 0.2143 & 0.2772 & 5.7172 & 0.3458 & 0.5546 \\
    \cdashline{2-7}
    & DSPO & 0.2131 & 0.2769 & 5.6825 & 0.3428 & 0.5642 \\
    & \hphantom{0} + \textbf{SDPO} & 0.2135 & 0.2777 & 5.6917 & 0.3441 & 0.5916 \\
    \cdashline{2-7}
    & DMPO$^\dagger$ & 0.2110 & 0.2710 & 5.5434 & 0.3382 & 0.2813 \\
    & \hphantom{0} + \textbf{SDPO} & \textbf{0.2144} & \textbf{0.2784} & \textbf{5.7312} & \textbf{0.3469} & \textbf{0.6369} \\
    \hline
    \multirow{11}{*}{HPS V2} 
    & SD 1.5 & 0.2088 & 0.2697 & 5.4933 & 0.3480 & -0.0469 \\
    & SFT & 0.2168 & 0.2838 & 5.7851 & 0.3591 & 0.6619 \\
    & Diff.-KTO & 0.2164 & 0.2766 & 5.6288 & 0.3420 & 0.5593 \\
    & MaPO$^\dagger$ & 0.2124 & 0.2760 & 5.6890 & 0.3528 & 0.3308 \\
    & DPOP$^\dagger$ & 0.2144 & 0.2780 & 5.7071 & 0.3563 & 0.3735 \\ 
    \cdashline{2-7}
    & Diff.-DPO & 0.2131 & 0.2743 & 5.6639 & 0.3552 & 0.1705 \\
    & \hphantom{0} + \textbf{SDPO}  & 0.2174 & 0.2827 & \bf 5.8744 & 0.3600 & 0.6211  \\
    \cdashline{2-7}
    & DSPO & 0.2168 & 0.2837 & 5.8346 & 0.3598 & 0.6483 \\
    & \hphantom{0} + \textbf{SDPO} & 0.2172 & 0.2847 & 5.8474 & 0.3586 & 0.6578\\
    \cdashline{2-7}
    & DMPO$^\dagger$ & 0.2131 & 0.2766 & 5.6538 & 0.3551 & 0.3171 \\
    & \hphantom{0} + \textbf{SDPO} & \textbf{0.2182} & \textbf{0.2848} & 5.8574 & \textbf{0.3612} & \textbf{0.7061} \\
    \hline
    \multirow{11}{*}{PartiPrompts} 
    & SD 1.5 & 0.2144 & 0.2724 & 5.3466 & 0.3343 & 0.0637 \\
    & SFT & 0.2181 & 0.2821 & 5.5981 & 0.3389 & 0.5830 \\
    & Diff.-KTO & 0.2178 & 0.2820 & 5.5630 & 0.3416 & 0.5697 \\
    & MaPO$^\dagger$ & 0.2152 & 0.2754 & 5.4754 & 0.3366 & 0.3358 \\
    & DPOP$^\dagger$ & 0.2169 & 0.2782 & 5.4894 & 0.3383 & 0.3644 \\ 
    \cdashline{2-7}
    & Diff.-DPO & 0.2167 & 0.2755 & 5.4045 & 0.3391 & 0.2560 \\
    & \hphantom{0} + \textbf{SDPO} & 0.2187 & 0.2815 & 5.5880 & 0.3423 & 0.5425  \\
    \cdashline{2-7}
    & DSPO & 0.2178 & 0.2819 & \bf 5.5997 & 0.3385 & 0.5640 \\
    & \hphantom{0} + \textbf{SDPO} & 0.2185 & \bf 0.2832 & 5.5975 & 0.3405 & 0.5955 \\
    \cdashline{2-7}
    & DMPO$^\dagger$ & 0.2163 & 0.2775 & 5.4724 & 0.3388 & 0.3653 \\
    & \hphantom{0} + \textbf{SDPO} & \bf 0.2190 & 0.2831 & 5.5956 & \bf 0.3430 & \bf 0.6381 \\
    \hline
\end{tabular}
\label{tab:score_15_full}
\end{table*}

\begin{table*}[t]
\centering
\footnotesize
\setlength{\tabcolsep}{3.5pt}
\caption{Full results of reward score comparison on Pick-a-Pic V2, HPS V2, and PartiPrompts using SDXL. $^\dagger$: results from our implementation due to the lack of official code.}
\begin{tabular}{l|l|ccccc}
    \hline
    \textbf{Dataset} & \textbf{Method} 
    & \textbf{PickScore($\uparrow$)} 
    & \textbf{\hphantom{0} HPS($\uparrow$) \hphantom{0}} 
    & \textbf{Aesthetics($\uparrow$)} 
    & \textbf{\hphantom{0} CLIP($\uparrow$) \hphantom{0}} 
    & \textbf{Image Reward($\uparrow$)} \\
    \hline
    \multirow{9}{*}{Pick V2} 
    & SDXL & 0.2242 & 0.2846 & 5.9970 & 0.3684 & 0.7382 \\
    & SFT & 0.2183 & 0.2809 & 5.7922 & 0.3658 & 0.5974 \\
    & MaPO & 0.2242 & 0.2871 & \bf 6.0979 & 0.3684 & 0.8359 \\
    \cdashline{2-7}
    & Diff.-DPO & 0.2251 & 0.2868 & 6.0115 & 0.3732 & 0.8357 \\
    & \hphantom{0} + \textbf{SDPO} & 0.2257 & 0.2876 & 5.9812 & 0.3746 & 0.8840 \\
    \cdashline{2-7}
    & DSPO & 0.2228 & 0.2834 & 5.8797 & 0.3756 & 0.8818 \\
    & \hphantom{0} + \textbf{SDPO} & 0.2240 & 0.2871 & 5.9529 & 0.3761 & 0.9238 \\
    \cdashline{2-7}
    & DMPO$^\dagger$ & 0.2253 & 0.2869 & 6.0119 & 0.3716 & 0.8555 \\
    & \hphantom{0} + \textbf{SDPO} & \bf 0.2263 & \bf 0.2882 & 5.9990 & \bf 0.3770 & \bf 0.9548 \\
    \hline
    \multirow{9}{*}{HPS V2} 
    & SDXL & 0.2290 & 0.2900 & 6.1271 & 0.3847 & 0.9047 \\
    & SFT & 0.2228 & 0.2883 & 5.9689 & 0.3806 & 0.8528 \\
    & MaPO & 0.2293 & 0.2934 & \bf 6.1882 & 0.3840 & 0.9703 \\
    \cdashline{2-7}
    & Diff.-DPO & 0.2288 & 0.2927 & 6.1380 & 0.3840 & 1.0159 \\
    & \hphantom{0} + \textbf{SDPO} & 0.2308 & 0.2938 & 6.1284 & 0.3879 & 1.0326 \\
    \cdashline{2-7}
    & DSPO & 0.2273 & 0.2916 & 6.0424 & 0.3894 & 1.0054 \\
    & \hphantom{0} + \textbf{SDPO} & 0.2293 & \bf 0.2944 & 6.1040 & 0.3889 & \bf 1.0745 \\
    \cdashline{2-7}
    & DMPO$^\dagger$ & 0.2302 & 0.2921 & 6.1101 & 0.3875 & 1.0154 \\
    & \hphantom{0} + \textbf{SDPO} & \bf 0.2308 & 0.2933 & 6.1113 & \bf 0.3897 & 1.0521 \\
    \hline
    \multirow{9}{*}{PartiPrompts} 
    & SDXL & 0.2277 & 0.2880 & 5.7901 & 0.3591 & 0.8573 \\
    & SFT  & 0.2221	& 0.2834 & 5.6496 & 0.3559 & 0.7515  \\
    & MaPO & 0.2278 & 0.2902 & \bf 5.8921 & 0.3580 & 0.9324 \\
    \cdashline{2-7}
    & Diff.-DPO & 0.2279 & 0.2900 & 5.8294 & 0.3629 & 1.0638 \\
    & \hphantom{0} + \textbf{SDPO} & 0.2290 & 0.2907 & 5.7882 & 0.3645 & 1.0654 \\
    \cdashline{2-7}
    & DSPO & 0.2261 & 0.2871 & 5.6947 & 0.3664 & 1.0514 \\
    & \hphantom{0} + \textbf{SDPO} & 0.2268 & 0.2897 & 5.7931 & \bf 0.3664 & \bf 1.1012 \\
    \cdashline{2-7}
    & DMPO$^\dagger$ & 0.2286 & 0.2904 & 5.8273 & 0.3610 & 0.9558 \\
    & \hphantom{0} + \textbf{SDPO} & \bf 0.2296 & \bf 0.2913 & 5.8103 & 0.3649 & 1.0623 \\
    \hline
\end{tabular}
\label{tab:score_xl_full}
\end{table*}

\begin{table*}
\centering
\small
\setlength{\tabcolsep}{5pt}
\caption{Average win rate comparison (\%) over the HPS V2 using SDXL.}
\begin{tabular}{l|l|cccccc}
\hline
\textbf{Model 1} & \textbf{Model 2} &
\textbf{Pick} & \textbf{HPS V2} & \textbf{Aesth.} & \textbf{CLIP} & \textbf{ImageReward} & \textbf{Mean} \\
\hline
\multicolumn{8}{l}{\textit{SDPO augmentation effect (base+SDPO vs base)}}\\
\hline
\makebox[4em][l]{Diff.-DPO} + SDPO    & Diff.-DPO & \bf 64.62 & 53.37 & 47.75 & \bf 57.75 & 51.88 & 55.08 \\
\makebox[4em][l]{DSPO}     + SDPO    & DSPO     & 64.25 & \bf 62.62 & \bf 59.75 & 48.62 & \bf 58.13 & \bf 58.67 \\
\makebox[4em][l]{DMPO}     + SDPO    & DMPO & 55.88 & 53.75 & 52.25 & 55.50 & 56.00 & 54.68 \\
\hline
\hline
\multicolumn{8}{l}{\textit{Versus SDXL}}\\
\hline
Diff.-DPO        & SDXL  & 48.25 & 63.38 & \bf 52.50 & 48.38 & 59.13 & 54.33  \\
\makebox[4em][l]{Diff.-DPO} + SDPO & SDXL  & 59.25 & 66.75 & 49.12 & 55.00 & 58.00 & 57.63 \\
DSPO            & SDXL  & 37.88 & 57.00 & 40.25 & 55.63 & 56.87 & 49.53 \\
\makebox[4em][l]{DSPO} + SDPO     & SDXL  & 54.63 & 68.00 & 44.62 & 55.00 & 65.12 & 57.47 \\
DMPO            & SDXL  & 58.63 & 64.88 & 46.00 & 53.62 & 62.00 & 57.03  \\
\makebox[4em][l]{DMPO} + SDPO     & SDXL & \bf 60.88 & \bf 68.50 & 47.63 & \bf 58.38  & \bf 68.63 & \bf 60.80 \\
\hline
\end{tabular}
\label{tab:hpsv2_winrate_sdxl}
\end{table*}

\end{document}